\pdfoutput=1
\documentclass[10pt,journal,compsoc]{IEEEtran}

\ifCLASSOPTIONcompsoc
\else
\fi
\ifCLASSINFOpdf
  \usepackage[pdftex]{graphicx}
\else
  \usepackage[dvips]{graphicx}
\fi
\usepackage[utf8]{inputenc}
\usepackage{multirow}
\usepackage[cmex10]{amsmath}
\usepackage[tight,footnotesize]{subfigure}

\usepackage{times}
\usepackage{url}
\usepackage{epsfig}
\usepackage{graphicx}
\usepackage{amssymb}
\usepackage{ifthen}
\usepackage{colortbl}
\usepackage{booktabs}
\usepackage{array}
\usepackage{bbold}
\usepackage{tikz}
\usepackage{pgfplots}
\usetikzlibrary{matrix,arrows,decorations.pathreplacing}
\DeclareMathOperator*{\argmax}{argmax}

\definecolor{cwblue1}{rgb}{0.27,0.427,0.623}
\definecolor{cwblue2}{rgb}{0.286,0.454,0.658}
\definecolor{cwblue3}{rgb}{0.733,0.811,0.905}

\pgfdeclarelayer{back}
\pgfsetlayers{back,main}

\makeatletter
\pgfkeys{%
  /tikz/on layer/.code={
    \def\tikz@path@do@at@end{\endpgfonlayer\endgroup\tikz@path@do@at@end}%
    \pgfonlayer{#1}\begingroup%
  }%
}
\makeatother

\newcommand{\symbm}{^{(m)}}
\newcommand{\symbk}{^{(k)}}
\newcommand{\symbn}{^{(n)}}

\begin{document}
\title{ModDrop: adaptive multi-modal\\ gesture recognition}

\author{Natalia Neverova,
        Christian Wolf,
        Graham Taylor and Florian Nebout% <-this % stops a space
\IEEEcompsocitemizethanks{\IEEEcompsocthanksitem N. Neverova and C. Wolf are with INSA-Lyon, LIRIS, UMR5205, F-69621, Universit\'e de Lyon, CNRS, France.\protect\\
E-mail: firstname.surname@liris.cnrs.fr
\IEEEcompsocthanksitem G. Taylor is with the School of Engineering, University of Guelph, Canada. %\protect\\
E-mail: gwtaylor@uoguelph.ca
\IEEEcompsocthanksitem F. Nebout is with Awabot, France. 
E-mail: florian.nebout@awabot.com
}% <-this % stops a space
\thanks{}}


\markboth{}%IEEE TRANSACTIONS ON PATTERN ANALYSIS AND MACHINE INTELLIGENCE}%
{Neverova \MakeLowercase{\textit{et al.}}: Multi-scale spatial and temporal integration for multi-modal gesture recognition}

%\IEEEpubid{\makebox[\columnwidth]{\hfill 0000--0000/00/\$00.00~\copyright~2007 IEEE}%

% PRIOR to the title within the \IEEEcompsoctitleabstractindextext IEEEtran

\IEEEcompsoctitleabstractindextext{%
\begin{abstract}

We present a method for gesture detection and localisation based on multi-scale and multi-modal deep learning. Each visual modality captures spatial information at a particular spatial scale (such as motion of the upper body or a hand), and the whole system operates at three temporal scales. Key to our technique is a training strategy which exploits: i) careful initialization of individual modalities; and ii) gradual fusion involving random dropping of separate channels (dubbed~\textit{ModDrop}) for learning cross-modality correlations while preserving uniqueness of each modality-specific representation. We present experiments on the \emph{ChaLearn 2014 Looking at People Challenge} gesture recognition track, in which we placed first out of 17 teams. Fusing multiple modalities at several spatial and temporal scales leads to a significant increase in recognition rates, allowing the model to compensate for errors of the individual classifiers as well as noise in the separate channels. Futhermore, the proposed \textit{ModDrop} training technique ensures robustness of the classifier to missing signals in one or several channels to produce meaningful predictions from any number of available modalities. In addition, we demonstrate the applicability of the proposed fusion scheme to modalities of arbitrary nature by experiments on the same dataset augmented with audio.

%\boldmath
%In the context of multi-modal gesture detection and recognition, we propose a deep recurrent architecture that iteratively learns and integrates discriminative data representations from individual channels, modeling cross-modality correlations and long-term temporal dependencies. Our framework integrates three data modalities: depth video, articulated pose and audio. Training recurrent architectures is known to be difficult, so in response we propose a novel algorithm for pre-training the architecture on individual modalities followed by an iterative representation fusing scheme.
%In our system, each gesture is decomposed into large-scale body motion and local subtle movements such as hand articulation. The idea of learning at multiple scales is also applied to the temporal dimension, such that a gesture is considered as an ordered set of characteristic motion impulses, or dynamic poses. %Each modality is first processed separately in short spatio-temporal blocks, where discriminative data-specific features are either manually extracted or learned. Finally, we employ a recurrent neural network for modeling large-scale temporal dependencies, data fusion and ultimately gesture classification.
%Our experiments on the 2013 Challenge on Multi-modal Gesture Recognition dataset have demonstrated state-of-the-art performance. Fusing multiple modalities at several spatial and temporal scales leads to a significant increase in recognition rates, allowing the model to compensate for errors of the individual classifiers as well as noise in the separate channels.
\end{abstract}
% not want either math or citations to appear in the abstract.

% Note that keywords are not normally used for peer review papers.
}
\maketitle
% is not selected <OR> if conference mode is selected - because compsoc
% conference papers position the abstract like regular (non-compsoc)
%\IEEEdisplaynotcompsoctitleabstractindextext
% \fi
%\IEEEpeerreviewmaketitle
%
%
\vspace*{-33pt}
%===========================================================
\section{Introduction}
\label{introduction}

\IEEEPARstart{G}{esture recognition} is one of the central problems in the rapidly growing fields of human-computer and human-robot interaction. Effective gesture detection and classification is challenging due to several factors: cultural and individual differences in tempos and styles of articulation, variable observation conditions, the small size of fingers in images taken in typical scenarios, noise in camera channels, infinitely many kinds of out-of-vocabulary motion, and real-time performance constraints.

Recently, the field of deep learning has made a tremendous impact in computer vision, demonstrating previously unattainable performance on the tasks of object detection and localization \cite{Girshick2014,Sermanet2014}, recognition \cite{Krizhevsky2012} and image segmentation \cite{Farabet2013,Couprie2014}. Convolutional neural networks (ConvNets) \cite{LeCun1998} have excelled on several scientific competitions such as ILSVRC \cite{Krizhevsky2012}, Emotion Recognition in the Wild
% (EmotiW 2013) 
\cite{Kahou2013}, Kaggle Dogs vs.~Cats \cite{Sermanet2014} and Galaxy Zoo. Taigman et al.~\cite{Taigman2014} recently claimed to have reached human-level performance using ConvNets for face recognition.
%Deep learning winning challenges: ImageNet \cite{Krizhevsky2012}, \cite{Sermanet2014}, emotions .  
%Deep learning in video classification: \cite{Karpathy2014}.
On the other hand, extending these models to problems involving the understanding of \textit{video} content is still in its infancy, this idea having been explored only in a small number of recent works \cite{Baccouche2012,Karpathy2014,Simonyan2014,modeep}. It can be partially explained by lack of sufficiently large datasets and the high cost of data labeling in many practical areas, as well as increased modeling complexity brought about by the additional temporal dimension and the interdependencies it implies \cite{zhou2013}.\looseness=-1
%difficulty of model training caused by introducing to the data an additional, different by nature temporal dimension.

The first gesture-oriented dataset containing a sufficient amount of training samples for deep learning methods was proposed for the \emph{ChaLearn 2013 Challenge on Multi-modal Gesture Recognition}.
%where a previous incarnation of our approach placed sixth in competition \cite{Neverova2013}. 
The deep learning method described in this paper placed first in the 2014 version of this competition \cite{escalera2014chalearn}.

\begin{figure}[!t]
\centering
\includegraphics[width=75mm]{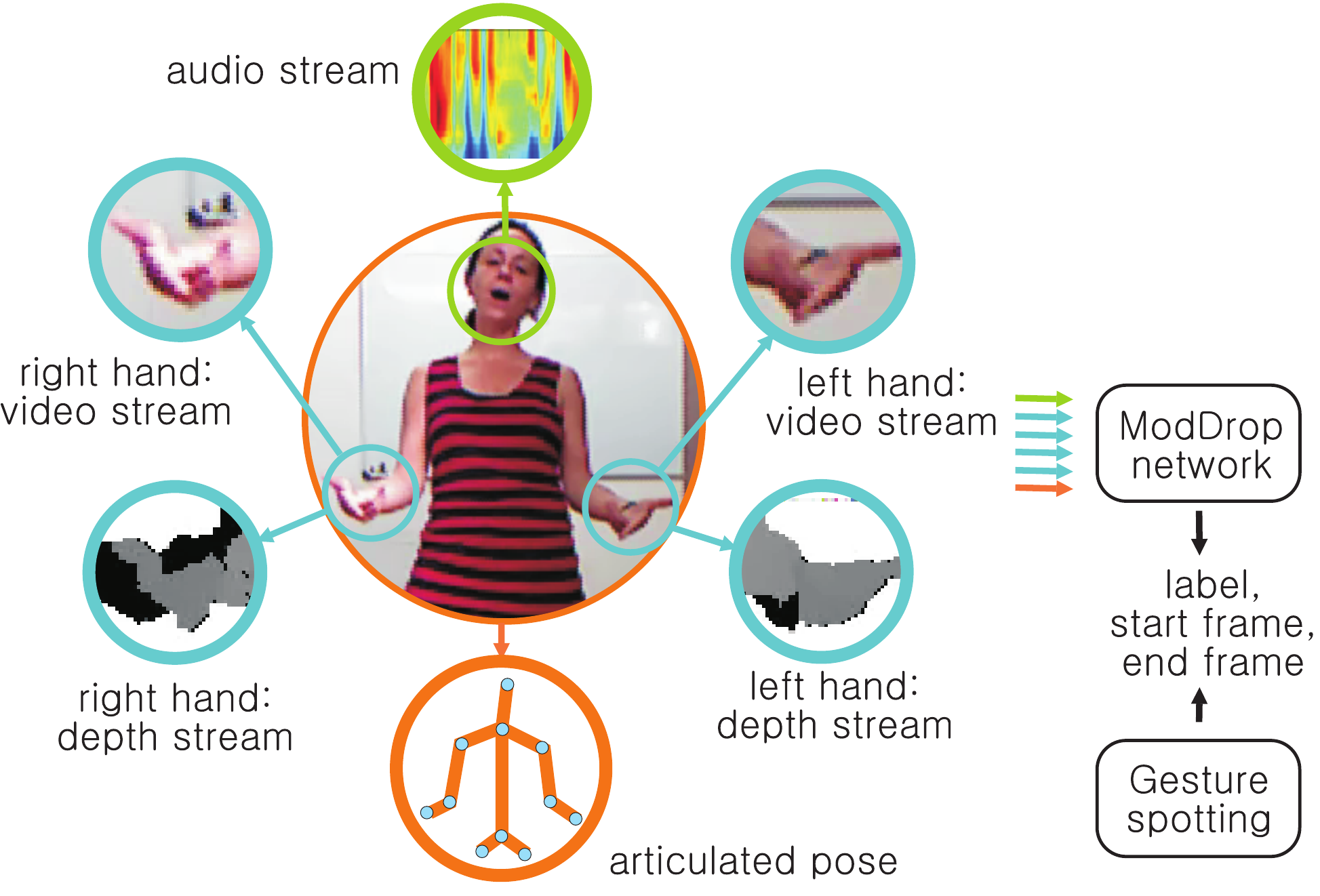}\vspace*{-5pt}
\caption{Overview of our method on an example from the 2014 ChaLearn Looking at People (LAP) dataset.} %Multi-modal and multi-scale gesture decomposition is followed by classification by the deep ModDrop network and is combined with parallel gesture localization}
\label{fig:teaser}\vspace*{-10pt}
\end{figure}

A core aspect of our approach is employing a multi-modal convolutional neural network for classification of so-called dynamic poses of varying duration (i.e.~temporal scales). %The best single scale configuration corresponding to a certain formulation of the dynamic pose alone places first %among challenge participants 
%(see Sec. \ref{sec:experiments_} for more details), while introducing parallel multi-scale paths leads to an additional gain in performance.  %Finally, we find it interesting to provide a comparison of the proposed approach with a baseline model employing a popular ensemble method. The performance of a hybrid solution, leading to another small gain, is reported for a reference. 
Visual data modalities integrated by our algorithm include intensity and depth video, as well as articulated pose information extracted from depth maps (see Fig. \ref{fig:teaser}).
% GWT articulated pose implies skeleton, I think 
We make use of different data channels to decompose each gesture at multiple scales not only temporally, but also spatially, to provide context for upper-body motion and more fine-grained hand/finger articulation. 

In this work, we pay special attention to developing an effective and efficient learning algorithm since learning large-scale multi-modal networks on a limited amount of labeled data is a formidable challenge. We also introduce an advanced training strategy, \textit{ModDrop}, that makes the network's predictions robust to missing or corrupted channels.% , parameter optimization is non-trivial.

%\begin{figure}[!t]
%\centering
%\includegraphics[width=\linewidth]{teaser.eps}
%\caption{Teaser!}
%\label{fig:teaser}
%\end{figure}

%CW REMOVED
%Our classifier outputs frame-wise prediction updates in real-time. Nevertheless, since temporal integration is involved, it suffers from a certain degree of inertia. Furthermore, due to high similarity between gesture classes on pre-stroke and post-stroke phases, frame-wise classification at those instants is often unreliable.
%To compensate for these negative effects, an additional module is introduced for filtering, denoising and more accurate gesture localization.

We demonstrate that the proposed scheme can be augmented with more data channels of arbitrary nature by introducing audio into the classification framework.

% GWT Let's remove this from the intro; it's not necessary (and you've already said "Finally!")
% In addition, we find it interesting to provide a comparison of the proposed approach with a baseline model employing a popular ensemble method. The performance of a hybrid solution %, leading to another small gain, 
% is reported for a reference. 

The major contributions of the present work are the following: We (i) develop a deep learning-based multi-modal and multi-scale framework for gesture detection, localization and recognition, which can be augmented with channels of an arbitrary nature (demonstrated by inclusion of audio); (ii) propose \textit{ModDrop} for effective fusion of multiple modality channels, which targets learning cross-modality correlations while prohibiting false co-adaptations between data representations and ensuring robustness of the classifier to missing signals; and (iii) introduce an audio-enhanced version of the ChaLearn 2014 LAP dataset.

%===========================================================
\section{Related work}
%
%\nn{Copied from JMLR, to edit (exactly the same as in ICCVW)}

%Motion related vision, which encompasses both gesture and action recognition, has gained the attention of many vision scientists over the last two decades. Dozens of papers published each year consider this problem in various contexts: still images, video, multiple views and range data, point clouds, etc. 
%
While having an immediate application in gesture recognition, this work addresses more general aspects of learning representations from raw data and multimodal fusion.

\subsubsection*{Gesture recognition}
Traditional approaches to action and distant gesture recognition from video typically include sparse or dense extraction of spatial or spatio-temporal engineered descriptors followed by classification \cite{Wang2013}.

Near-range applications may require more accurate reconstruction of hand shapes. %In this case, fitting a 3D hand model, as well as appearance-based algorithms provide more appropriate solutions. 
A group of recent works is dedicated to inferring the hand pose through pixel-wise hand segmentation and estimating the positions of hand or body joints %in a bottom-up fashion
 \cite{Shotton2011,Keskin2011,Tang2013,Tang2014,Tompson2014,neverovaACCV2014}, 
tracking 
%In parallel, tracking-based approaches are advancing quickly 
\cite{Oikonomidis2011,Qian2014}
%. On the other hand, in \cite{Tang2014} the authors proposed the Latent Regression Forest for coarse-to-fine search of joint positions.
and graphical models 
%Finally, graphical models, exploring spatial relationships between body and hand parts, have recently attracted close attention 
\cite{Wang2013a,chen2014}.

Multi-modal aspects are of relevance in this domain. In \cite{Wang2012}, a combination of skeletal features and local occupancy patterns (LOP) were calculated from depth maps to describe hand joints. In
\cite{Sung2012}, skeletal information was integrated in two ways for extracting HoG features from RGB and depth images: either from global bounding boxes containing a whole body or from regions containing an arm, a torso and a head.
Similarly, \cite{Chen2013, chalearn2, chalearn3} fused skeletal information with HoG features extracted from either RGB or depth, while \cite{Nandakumar2013} proposed a combination of a covariance descriptor representing skeletal joint data with spatio-temporal interest points extracted from RGB augmented with audio.

%\vspace*{-5mm}
%\subsubsection*{Representation learning}
Various multi-layer architectures have been proposed in the context of motion analysis for \textit{learning} (as opposed to handcrafting) representations directly from data, either in a supervised or unsupervised way. % are based on minimizing reconstruction error. %, e.g.~as in autoencoders, or some predefined energy function, e.g.~as in restricted Boltzmann machines (RBMs). 
Independent subspace analysis (ISA) \cite{Le2011a} as well as autoencoders \cite{Ranzato2007,Baccouche2012} are examples of efficient unsupervised methods for learning  hierarchies of invariant spatio-temporal features. %A similar by spirit group of autoencoders \cite{Baccouche2012} adapted the original work of \cite{Ranzato2007} by adding a temporal dimension to 2D sparse convolutional autoencoders. %\cite{Taylor2010} extended and scaled up the gated RBM (GRBM) model proposed by~\cite{Memisevic2007} for learning representations of image patch transformations. f
Space-time deep belief networks \cite{Chen2010a} produce high-level representations of video sequences using convolutional RBMs. 

Vanilla supervised convolutional networks have also been explored in this context. A method proposed in \cite{Ji2013} 
is based on low-level preprocessing of the video input and employs a 3D convolutional network for learning of mid-level spatio-temporal representations and classification. Pigou et al. \cite{chalearn5} explored this approach in the context of sign language recognition from depth video, while Wu and Chao \cite{chalearn6}  employed a combination of convnets with HMMs.
Recently, Karpathy et al.~\cite{Karpathy2014} have proposed a convolutional architecture for large-scale video classification operating at two spatial resolutions (fovea and context streams).

In contrast to existing solutions, in this work we propose a novel specific tree-structured deep learning architecture allowing to classify hand gestures with higher accuracy while restricting the number of free parameters.

\begin{figure*}[!t]
\centering
\includegraphics[width=120mm]{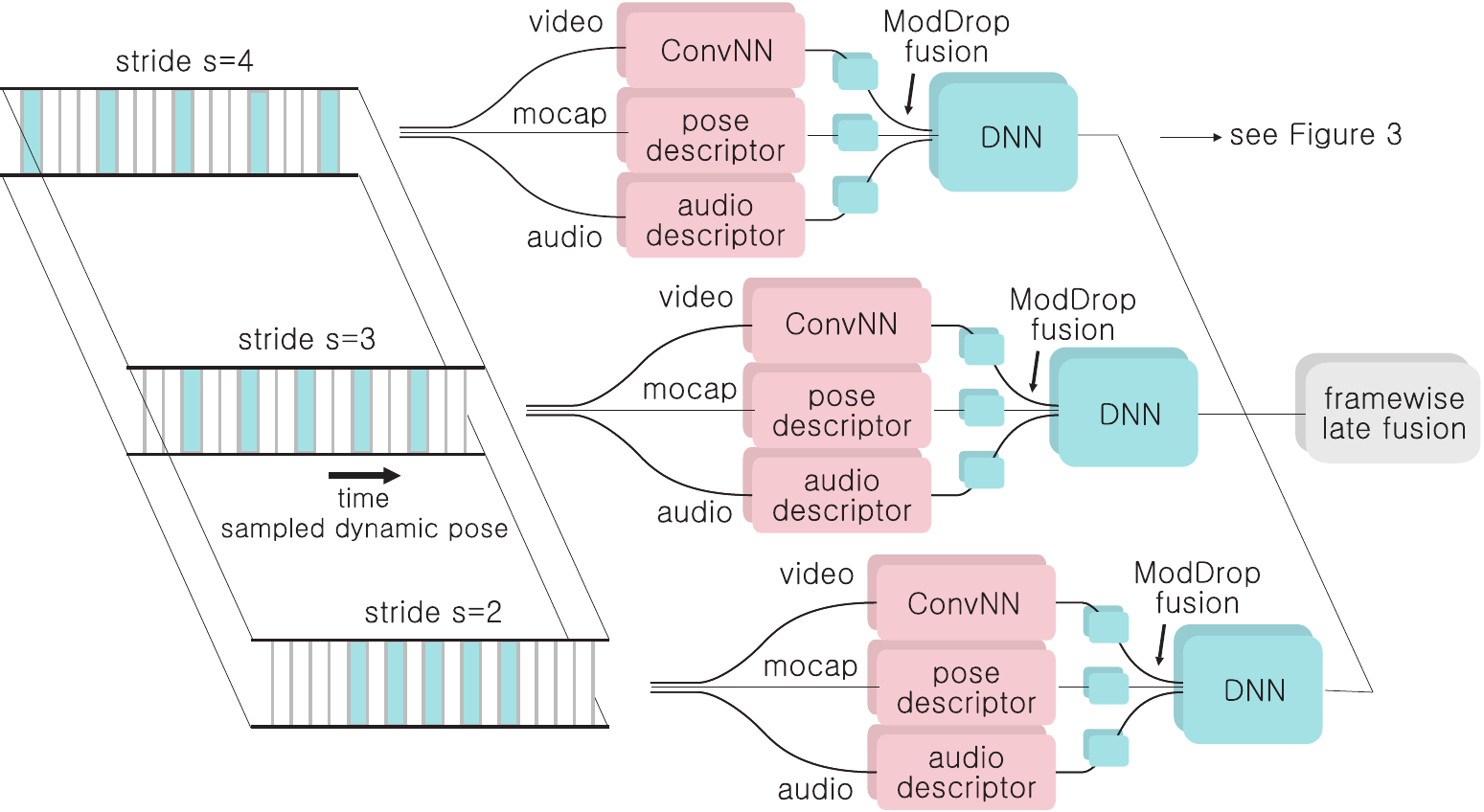}
\centering\caption{\hspace*{\fill}The ModDrop network operating at 3 temporal scales corresponding to 3 durations of dynamic poses.\hspace*{\fill}}%corresponding to dynamic poses of 3 durations.} %Although the audio modality is not present in the ChaLearn 2014 LAP dataset, we present additional experiments by augmenting the visual signal with audio recordings from the 2013 version of the data.}
\label{fig:overview}
\vspace*{-10pt}
\end{figure*}

\subsubsection*{Multi-modal fusion}
While in most practical applications, late fusion of scores output by several models offers a cheap and surprisingly effective solution \cite{Kahou2013}, both late and early fusion of either final or intermediate data representations remain under active investigation. 

A significant amount of work on early combining of diverse feature types has been applied to object and action recognition. Multiple Kernel Learning (MKL) \cite{mkl2004} has been actively discussed in this context. At the same time, as shown by \cite{gehler2009}, simple additive or multiplicative averaging of kernels may reach the same level of performance while being orders of magnitude faster.

Ye et al.~\cite{ye2012} proposed a late fusion strategy compensating for errors of individual classifiers by minimising the rank of a score matrix. %In follow-up work \cite{liu2013}, they identified sample-specific optimal fusion weights by enforcing similarity in fusion scores for visually similar labeled and unlabeled samples. Xu et al.~\cite{xu2013} introduced the Feature Weighting via Optimal Thresholding (FWOT) algorithm jointly optimising feature weights and thresholds.
Nataranjan et al.~\cite{natarajan2012} employed multiple strategies, including MKL-based combinations of features, Bayesian model combination, and weighted average fusion of scores from multiple systems.

A number of deep architectures have recently been proposed specifically for multi-modal data. Ngiam et al.~\cite{Ngiam2011} employed sparse RBMs and bimodal deep antoencoders to learn cross-modality correlations in the context of audio-visual speech classification of isolated letters and digits. Srivastava et al.~\cite{Srivastava2013} used a multi-modal deep Boltzmann machine in a generative fashion to tackle the problem of integrating images and text annotations. Kahou et al.~\cite{Kahou2013} won the 2013 Emotion Recognition in the Wild Challenge by training convolutional architectures on several modalities, such as facial expressions from video, audio, scene context and features extracted around mouth regions.
Wu et al. \cite{wu2014mm} paid special attention to exploring inter-feature and inter-class relationships in deep neural networks for video analysis. Finally, in~\cite{Neverova2013} the authors proposed a multi-modal convolutional network for gesture detection and classification from a combination of depth, skeletal information and audio. 

In this work, we explore multimodal deep learning in more detail and pay special attention to encorporating the specifics of multimodality in the training procedure.

\section{Gesture classification}

On a dataset such as \emph{ChaLearn 2014 LAP}, we face several key challenges: learning representations at multiple spatial and temporal scales, integrating the various modalities, and training a complex model when the number of labeled examples is not at \emph{web-scale} like static image datasets (e.g.~\cite{Krizhevsky2012}). We start by describing how the first two challenges are overcome at an architectural level. Our training strategy addressing the last issue is described in Sec.~\ref{sec:training}.

Our proposed multi-scale deep neural network consists of a combination of single-scale paths connected in parallel (see Fig.~\ref{fig:overview}). Each path independently learns a representation and performs gesture classification at its own temporal scale given input from RGBD video and pose signals (an audio channel can be also added, if available). Predictions from all paths are aggregated through additive late fusion. 

%CW removed : we do not really have long term strategies anyway
%This strategy aims to first extract the most salient (discriminative) motions at a fine temporal resolution and, at the same time, consider them in the context of global gesture structure, smoothing and compensating for per-block errors typical for a given gesture class.

%, first, automatic convolutional learning data representations from video and, second, handcrafted local descriptors. 
%Fig.~\ref{fig:overview} illustrates the fusion of outputs of a temporally multiscale deep learning based architecture and an alternative ensemble method. 

To differentiate among temporal scales, a notion of \textit{dynamic pose} is introduced, meaning a sequence of video frames, synchronized across modalities, sampled with a given temporal stride $s$ and concatenated to form a spatio-temporal 3D volume (similar to earlier works, such as \cite{hernandezvela2014}). Varying the value of $s$ allows the model to leverage multiple temporal scales for prediction, accommodating differences in tempos and styles of articulation. 
%of different users. 
Our model is therefore different from the one proposed in \cite{Farabet2013}, where by ``multi-scale'' Farabet et al.~imply a multi-resolution spatial pyramid rather than a fusion of temporal sampling strategies.
% GWT: removing "allows the model to switch from one temporal scale to another" and notion of "duty cycle" because there's no discrete notion of switching as far as I can tell; fusion is blending them in a smooth way but with fixed windows, right?
%rather than a duty cycle of temporal sampling.
Regardless of the stride $s$, we use the same number of frames (5) at each scale.
Fig.~\ref{fig:overview} shows the paths used in this work.
% (with $s=2{\dots}4$). 
At each scale and for each dynamic pose, the classifier outputs a per-class score.
%, while the last path represents a baseline operating at a single scale ($s=4$).
%GWT: remove probability, replace with score
%single probability distribution over classes

\begin{figure*}[!t]
	\begin{center}
		\includegraphics[width=0.72\linewidth]{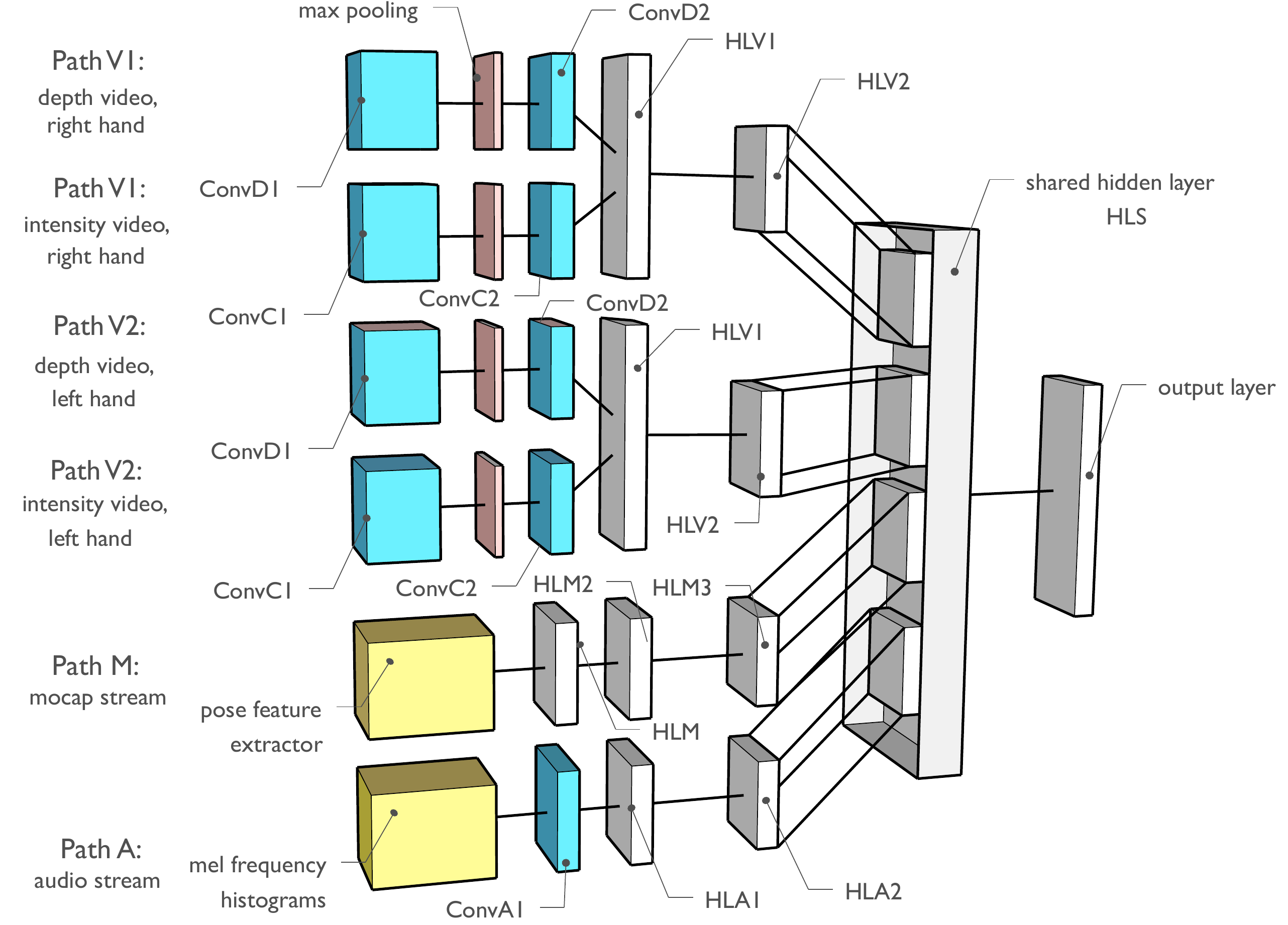}
	\end{center}\vspace*{-12pt}
   	\caption{Single-scale deep architecture. Individual classifiers are pre-trained for each data modality (paths V1, V2, M, A) and then fused using a 2-layer fully connected network initialized in a specific way (see Sec. \ref{sec:training}).}% The first layers of paths V1 and V2 perform 3D convolutions followed by 3D max pooling shrinking the temporal dimension. The second layers on these paths are exclusively spatial. Weights are shared across V1 and V2 path.}
	\label{fig:architecture}\vspace*{-8pt}
\end{figure*}

All available modalities, such as depth, gray scale video,  articulated pose, and eventually audio, contribute to the network's prediction. Global appearance of each gesture instance is captured by the skeleton descriptor, while video streams convey additional information about hand shapes and their dynamics which are crucial for discriminating between gesture classes performed in similar body poses. 

Due to the high dimensionality of the data and the non-linear nature of cross-modality structure, an immediate concatenation of raw skeleton and video signals is sub-optimal. However, initial discriminative learning of individual data representations from each isolated channel followed by fusion has proven to be efficient in similar tasks \cite{Ngiam2011}. Therefore, we first learn discriminative data representations within each separate channel, followed by joint fine tuning and fusion by a meta-classifier independently at each scale. More details are given in Sec.~\ref{sec:training}. A shared set of hidden layers is employed at different levels for, first, fusing of ``similar by nature'' gray scale and depth video streams and, second, combining the obtained joint video representation with the transformed articulated pose descriptor (and audio signal, if available).\vspace*{-2pt}

\subsection{Articulated pose}
\label{sec:skeleton}

%\begin{figure}[!t]
%	\begin{center}
%		\includegraphics[width=80mm]{ladies_light_contours.eps}
%	\end{center}
%   	\caption{The pose descriptor is calculated from normalized coordinates of 11 upper body joints (on the left) also including their velocities and accelerations, a set of angles (triples of joints forming inclination angles are shown on the right) and pairwise distances. The body coordinate system (shown in blue on the left) is calculated from 6 torso joints (shown in dark gray on the left).}
%	\label{fig:posedescriptor}
%\end{figure}

The full body skeleton provided by modern consumer depth cameras and associated middleware consists of 20 or fewer joints identified by their coordinates in a 3D coordinate system aligned with the depth sensor. For our purposes we exploit only 11 joints corresponding to the upper body: Head, Shoulder and Hips central points, as well as left and right Hip, Shoulder, Elbow and Hand joints. % (see Fig.~\ref{fig:posedescriptor}). 
%We do not use wrist joints as their positions are often unstable.

We formulate a pose descriptor consisting of 7 logical subsets as described in \cite{neverovaECCVW2014}. %, , and allow the classifier to perform online feature selection.
%Raw, i.e.~pre-normalization, positions of 11 upper body joints in a 3D coordinate system associated with the depth sensor are denoted as $\mathbf{p}^{(i)}_{\textrm{raw}}{=}\{x^{(i)},y^{(i)},z^{(i)}\}$, $i{=}0...10$ ($i{=}0$ corresponds to the \emph{HipCenter} joint).
%
Following~\cite{zanfir2013movingpose}, we first calculate normalized joint positions, as well as their velocities and accelerations, and then augment the descriptor with a set of characteristic angles and pairwise distances.

The skeleton is represented as a tree structure with the \emph{HipCenter} joint playing the role of a root node. Its coordinates are subtracted from the rest of the vectors 
%$\mathbf{p}_{\textrm{raw}}$ 
to eliminate the influence of position of the body in space. To compensate for differences in body sizes, proportions and shapes, we start from the top of the tree and iteratively normalize each skeleton segment to a corresponding average ``bone" length estimated from all available training data. %It is done in the way that absolute joint positions are corrected while corresponding orientations remain unchanged:\vspace*{-5pt}
%\begin{align}
%\mathbf{p}^{(i)}(t) = \mathbf{p}_{\textrm{raw}}^{(i-1)}(t) + \dfrac{\mathbf{p}^{(i)}_{\textrm{raw}}(t)-\mathbf{p}_{\textrm{raw}}^{(i-1)}(t)}{||\mathbf{p}^{(i)}_{\textrm{raw}}(t)-\mathbf{p}^{(i-1)}_{\textrm{raw}}(t)||}b^{(i-1,i)} - \mathbf{p}^{(0)}_{\textrm{raw}}(t),
%\end{align}
%\begin{align}
%\mathbf{p}^{(i)} = \mathbf{p}_{\textrm{raw}}^{(i-1)} + \dfrac{\mathbf{p}^{(i)}_{\textrm{raw}}-\mathbf{p}_{\textrm{raw}}^{(i-1)}}{||\mathbf{p}^{(i)}_{\textrm{raw}}-\mathbf{p}^{(i-1)}_{\textrm{raw}}||}b^{(i-1,i)} - \mathbf{p}^{(0)}_{\textrm{raw}},
%\end{align}\vspace*{-8pt}\\
%where  $\mathbf{p}^{(i)}_{\textrm{raw}}$ is a  current joint, $\mathbf{p}^{(i{-}1)}_{\textrm{raw}}$ is its direct predecessor in the tree, $b^{(i-1,i)}$, $i=1{\ldots}10$ is a set of estimated average lengths of ``bones" and $\mathbf{p}$ are normalized joints. 
Once the normalized \textit{joint positions} are obtained, we perform Gaussian smoothing along the temporal dimension ($\sigma{=}1$, filter $5{\times}1$) to decrease the influence of skeleton jitter.

\textit{Joint velocities} and \textit{joint accelerations} are calculated as first and second derivatives of normalized joint positions.%:
%\begin{align}
%$\delta \mathbf{p}^{(i)}(t) \approx \mathbf{p}^{(i)}(t+1)-\mathbf{p}^{(i)}(t-1)$.
%\end{align}
%
%\textbf{Joint accelerations} are the second derivatives of the same positions.%: 
%\begin{align}
%$\delta^2 \mathbf{p}^{(i)}(t)\!\approx\!\mathbf{p}^{(i)}\!(t+2)\! + \!\mathbf{p}^{(i)}\!(t-2)\! - \!2\mathbf{p}^{(i)}\!(t).$
%\end{align}

\textit{Inclination angles} are formed by all triples of anatomically connected joints %$(i,j,k)$, 
plus two ``virtual'' angles~\cite{neverovaECCVW2014}.% (Right,Left)\emph{Elbow}-(Right,Left)\emph{Hand}-\emph{HipCenter}.%:\vspace*{-5pt} %(Fig. \ref{fig:posedescriptor}):
%
%\begin{align}
%\alpha^{(i,j,k)}= \arccos{\dfrac{(\mathbf{p}^{(k)}-\mathbf{p}^{(j)})(\mathbf{p}^{(i)}-\mathbf{p}^{(j)})}{||\mathbf{p}^{(k)}-\mathbf{p}^{(j)}||\cdot ||\mathbf{p}^{(i)}-\mathbf{p}^{(j)}||}}
%\end{align}\vspace*{-12pt}

\textit{Azimuth angles} provide additional information about the pose in the coordinate space associated with the body. We apply PCA on the positions of 6 torso joints. % (\emph{HipCenter}, \emph{HipLeft}, \emph{HipRight}, \emph{ShoulderCenter}, \emph{ShoulderLeft}, \emph{ShoulderRight}) %(Fig. \ref{fig:posedescriptor}) 
%
%to obtain 3 vectors forming the basis: $\left\{\mathbf{u}_x,\mathbf{u}_y, \mathbf{u}_z\right\}$, where $\mathbf{u}_x$ is approximately parallel to the shoulder line, $\mathbf{u}_y$ is aligned with the spine and $\mathbf{u}_z$ is perpendicular to the torso.
Then for each pair of connected bones, we calculate angles between projections of the second bone %($\mathbf{v}_2$) 
and the vector %$\mathbf{u}_x$ ($\mathbf{v}_1$) 
on the plane perpendicular to the orientation of the first bone. %As in the previous case, we also include two virtual ``bones'' (Right,Left)\emph{Hand}-\emph{HipCenter}.\vspace{-5pt}
%\begin{equation*}
%\mathbf{v}_1 = \mathbf{u}_x - (\mathbf{p}^{(j)}-\mathbf{p}^{(i)})\dfrac{\mathbf{u}_x\cdot(\mathbf{p}^{(j)}-\mathbf{p}^{(i)})}{||\mathbf{p}^{(j)}-\mathbf{p}^{(i)}||^2}
%\end{equation*}\vspace{-5pt}
%\begin{equation*}
%\mathbf{v}_2\! =\!(\mathbf{p}^{(k)}\!-\!\mathbf{p}^{(j)})\! -\! (\mathbf{p}^{(j)}\!-\!\mathbf{p}^{(i)})\dfrac{(\mathbf{p}^{(k)}\!-\!\mathbf{p}^{(j)})(\mathbf{p}^{(j)}\!-\!\mathbf{p}^{(i)})}{||\mathbf{p}^{(j)}-\!\mathbf{p}^{(i)}||^2}
%\end{equation*}\vspace{-5pt}
%\begin{equation}
%\beta^{(i,j,k)} = \arccos{\dfrac{\mathbf{v}_1\cdot\mathbf{v}_2}{||\mathbf{v}_1||||\mathbf{v}_1||}}
%\end{equation}\vspace*{-12pt}\\

\textit{Bending angles} are a set of angles between a basis vector, %$\mathbf{u}_{z}$, 
perpendicular to the torso, and joint positions.%:\vspace*{-5pt}
%\begin{align}
%\gamma^{(i)} = \arccos{\dfrac{\mathbf{u}_z\cdot\mathbf{p}^{(i)}}{||\mathbf{p}^{(i)}||}}
%\end{align}\vspace*{-12pt}\\

Finally, we include \textit{pairwise distances} between all normalized joint positions.%:
%\begin{equation}
%$\rho^{(i,j)}{=}||\mathbf{p}^{(i)}_n - \mathbf{p}^{(j)}_n||$.
%\end{equation}

Combined together, this produces a 183-dimensional pose descriptor for each video frame.
%: 
%$
%	\mathbf{D}{=}[\mathbf{p}, \delta\mathbf{p}, \delta^2\mathbf{p}, \boldsymbol{\alpha}, \boldsymbol{\beta}, \boldsymbol{\gamma}, \boldsymbol{\rho}]^T.		
%$
Finally, each feature is normalized to zero mean and unit variance.

A set of consequent 5 frame descriptors sampled with a given stride $s$ are concatenated to form a $915$-dimensional dynamic pose descriptor which is further used for gesture classification.
The two subsets of features involving derivatives contain dynamic information and for dense sampling may be partially redundant as several occurrences of the same frame are stacked when a dynamic pose descriptor is formulated. Although theoretically unnecessary, this is beneficial when the amount of training data is limited.\vspace*{-2pt}

\subsection{Depth and intensity video}%: convolutional learning}
\label{videofeatures}

%\begin{figure}[!t]
%\centering\hspace*{10pt}
%\includegraphics[width=0.97\linewidth]{network.pdf}%eccv2014w_network_}
%\caption{Single-scale deep architecture. Individual classifiers are pre-trained for each data modality (paths V1, V2, M) and then fused using a 2-layer shared fully connected network initialized in a specific way (see Sec. \ref{training}). The first layers perform 3D convolutions followed by 3D max pooling shrinking the temporal dimension. The second layers are exclusively spatial. Weights are shared across V1 and V2 paths.}
%\label{fig:network}
%\end{figure}

Two video streams serve as a source of information about hand pose and finger articulation. Bounding boxes containing images of hands are cropped around positions of the \emph{RightHand} and \emph{LeftHand} joints. 
To eliminate the influence of the person's position with respect to the camera and keep the hand size approximately constant, the size of each bounding box is normalized by the distance between the hand and the sensor.%:\vspace*{-3pt}
%\begin{equation}
% H_x = \dfrac{h_xX}{z \cdot \tan{(\alpha_{FoV,x})}},\quad
% H_y = \dfrac{h_yY}{z \cdot \tan{(\alpha_{FoV,y})}},
%\end{equation}\vspace*{-12pt}\\
%where $H_{x}$ and $H_{y}$ are the hand sizes (px) along the x and y axes of the camera sensor, $h_{x}$ and $h_{y}$ are the physical sizes of an average hand, in mm, $X, Y$ are the frame dimensions, in pixels, $z$ is the distance between the hand and the camera in mm, and $\alpha_{\text{FoV},x}$ and $\alpha_{\text{FoV},y}$ are the camera field of view along the x and y axes respectively.

Within each set of frames forming a dynamic pose, hand position is stabilized by minimizing inter-frame square-root distances calculated as a sum over all pixels, and corresponding frames are concatenated to form a single spatio-temporal volume.
The color stream is converted to gray scale, and both depth and intensity frames are normalized to zero mean and unit variance. Left hand videos are flipped about the vertical axis and combined with right hand instances in a single training set.

During modality-wise pre-training, video pathways are adapted to produce predictions for each hand, rather than for the whole gesture. Therefore, we introduce an additional step to eliminate possible noise associated with switching from one active hand to another. For one-handed gesture classes, we detect the active hand and adjust the class label for the inactive one. In particular, we estimate the motion trajectory length of each hand using the respective joints provided by the skeleton stream (summing lengths of hand trajectories projected to the $x$ and $y$ axes):\vspace*{-7pt}
\begin{align}
\Delta = \sum_{t=2}^5(|x(t)-x(t-1)| + |y(t)-y(t-1)|),
\end{align}\vspace*{-10pt}\\
\noindent where $x(t)$ is the x-coordinate of a hand joint (either left or right) and $y(t)$ is its y-coordinate.
Finally, the hand with a greater value of $\Delta$ is assigned the label class, while the other hand is assigned the zero-class ``no action'' label. 

For each channel and each hand, we perform 2-stage convolutional learning of data representations independently (first in 3D, then in 2D, see Fig.~\ref{fig:architecture}) and fuse the two streams with a set of fully connected hidden layers. Parameters of the convolutional and fully-connected layers at this step are shared between the right hand and left hand pathways. 
Our experiments have demonstrated that relatively early fusion of depth and intensity features leads to a significant increase in performance, even though the quality of predictions obtained from each channel alone is unsatisfactory.

\vspace*{-5pt}\subsection{Audio stream}
\label{sec:audio}

Recent advances in the field of speech processing have demonstrated that using weakly preprocessed raw audio data in combination with deep learning leads to higher performance relative to state-of-the-art systems based on hand crafted features (typically from the family of Mel-frequency cepstral coefficients, or MFCC).
Deng et al.~\cite{Deng2013} demonstrated the advantage of using primitive spectral features, such as 2D spectrograms, in combination with deep autoencoders. 
Ngiam et al.~\cite{Ngiam2011} applied the same strategy to the task of multi-modal speech recognition while augmenting the audio signal with visual features. Further experiments from Microsoft \cite{Deng2013} have shown that ConvNets appear to be especially efficient in this context since they allow the capture and modeling of structure and invariances that are typical for speech. 

Comparative analysis of our previous approach \cite{Neverova2013} based on phoneme recognition from sequences of MFCC features and a deep learning framework has demonstrated that the latter strategy allows us to obtain significantly better performance on the ChaLearn dataset (see Sec.~\ref{sec:experiments_} for more details). Therefore, in this work, the audio signal is processed in the same manner as video data, i.e.~by feature learning within a convolutional architecture. 

\begin{figure}[!t]
%\begin{subfigure}[b]{0.48\linewidth}
	\begin{center}
		\includegraphics[width = 20mm,height=20mm]{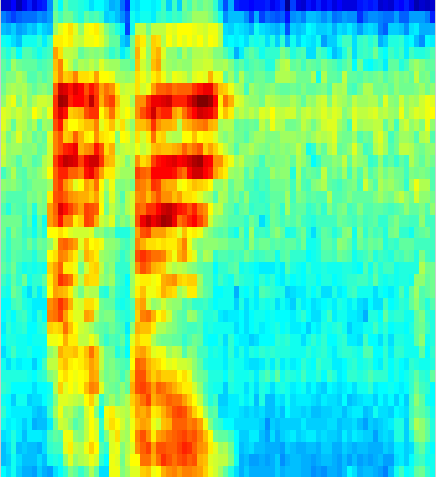}\hspace*{1pt}
		\includegraphics[width = 20mm,height=20mm]{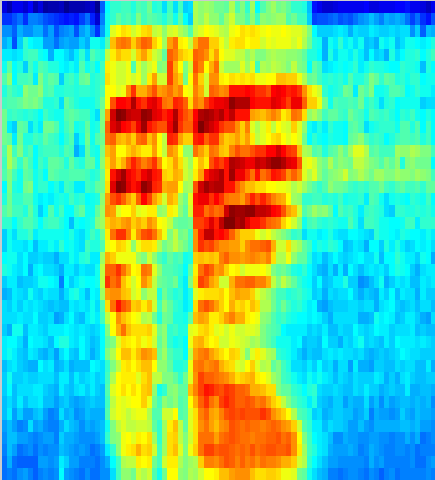}\hspace*{3pt}
	%\end{center}
%	\label{fig:cosatifarei}
%	\caption{``cosa ti farei''}
%\end{subfigure}\hspace*{\fill}
%\begin{subfigure}[b]{0.48\linewidth}
	%\begin{center}
	\includegraphics[width = 20mm,height=20mm]{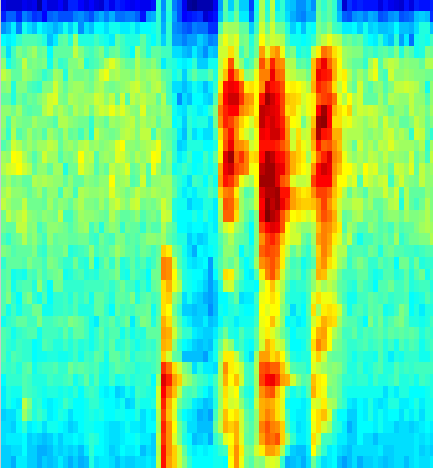}\hspace*{1pt}
	\includegraphics[width = 20mm,height=20mm]{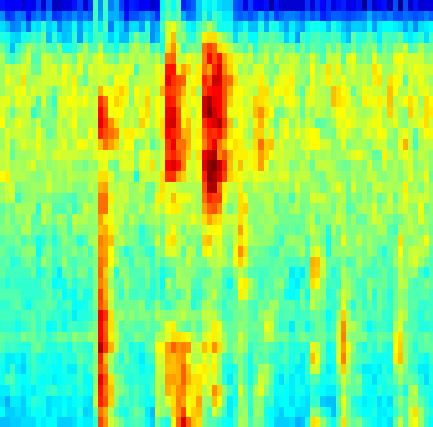}
	\end{center}
%	\label{fig:perfetto}
%	\caption{``perfetto''}
%\end{subfigure}
\vspace*{-8pt}
   	\caption{Mel-scaled spectrograms of two pairs of audio samples corresponding to two different gestures.}%types of Italian conversational gestures (``cosa ti farei" and ``perfetto'').}% (shown in false colors).}
	\label{fig:audio}\vspace*{-10pt}
\end{figure}

To preprocess, we perform basic noise filtering and speech detection by thresholding the raw signal along the absolute value of the amplitude~($\tau_1$). Short, isolated peaks of duration less than~$\tau_2$ are also ignored during training. 
We apply a short-time Fourier transform on the raw audio signal to obtain a 2D local spectrogram which is further transformed to the Mel	-scale to produce 40 log filter-banks on the frequency range from 133.3 to 6855.5 Hz, i.e.~the zero-frequency component is eliminated. 
In order to synchronize the audio and visual signals, the size of the Hamming window is chosen to correspond to the duration of~$L_1$ frames with half-frame overlap. A typical output is illustrated in Fig. \ref{fig:audio}. As it was experimentally demonstrated by \cite{Deng2013}, the step of the scale transform is important. Even state-of-the-art deep architectures have difficulty learning these kind of non-linear transformations.

%To obtain additional training data, audio sequences are sampled at sub-frame frequency as previously described for the articulated pose. All spectrograms are then normalized to the unit range.

A one-layer convolutional network in combination with two fully-connected layers form the corresponding path which we, as before, pretrain for preliminary gesture classification from short utterances. The output of the penultimate layer provides audio features for data fusion and modeling temporal dependencies (see Sec.~\ref{sec:training}). 

%% General system description

%%% Local Variables: 
%%% mode: latex
%%% TeX-master: "main"
%%% End: 
\vspace*{-5pt}

%%% Local Variables: 
%%% mode: latex
%%% TeX-master: "main"
%%% End: 

\section{Training procedure}
\label{sec:training}
%Although for many state-of-the-art deep learning networks selection of hyper-parameters is up to a certain degree ad-hoc or arbitrary, appropriate choice of an architecture and a training algorithm is a key to high performance in practice. 
 In this section we describe the most important architectural solutions that were critical for our multi-modal setting: per-modality pre-training and aspects of fusion such as the initialization of shared layers. Also, we introduce the concept of multi-modal dropout (ModDrop), which makes the network less sensitive to loss of one or more channels.\vspace*{-2pt}

\subsubsection*{Pretraining} Depending on the source and physical nature of a signal, input representation of any modality is characterized by its dimensionality, information density, and associated correlated and uncorrelated noise. %, systematic errors in measurements.  
Accordingly, a monolithic network taking as an input a combined collection of features from all channels is suboptimal, since a uniform distribution of parameters over the input is likely to overfit one subset of features and underfit the others. Here, performance-based optimization of hyper-parameters may resolve in cumbersome architectures requiring sufficiently larger amounts of training data and computational resources at training and test times. Furthermore, blind fusion of fundamentally different signals at early stages has a high risk of learning false cross-modality correlations and dependencies among them (see Sec.~\ref{sec:experiments_}). %\gwt{It feels like there should be at least one citation here.}
%\nn{Not sure which one, it is just an intuition.}
%\cw{Not sure ... perhaps the Ng paper (audio/video) will do. We will need to reread it.}
To capture complexity within each channel, separate pretraining of input layers and optimization of hyper parameters for each subtask are required. %In this work, for each data channel we employ early stopping on different subsets of the validation set to prevent overfitting of both training and validation data. %\gwt{I've re-read this last sentence many times and I don't understand the concept here about using different subsets of the validation data for different channels.} \nn{Since we optimize different parts of the network separately and use early stopping, probably we don't want to tune each of them to the exact set of validation samples? On MNIST it increases the difference between validation and test errors.}
%\cw{I agree with Graham in that it is not clear. I would understand temporally switching between different validation sets for the same channel, but not over channels, since they are independant at this stage.}

Recall Fig.~\ref{fig:architecture} illustrating a single-scale deep multi-modal convolutional network. Initially it starts with six separate pathways: depth and intensity video channels for right (V1) and left (V2) hands, a mocap stream (M) and an audio stream (A).  From our observations, inter-modality fusion is effective at early stages if both channels are of the same nature and convey complementary information. On the other hand, mixing modalities which are weakly correlated, is rarely beneficial until the final stage. Accordingly, in our architecture, two video channels corresponding to each hand (layers HLV1 and HLV2) are fused immediately after feature extraction. We postpone any attempt to capture cross-modality correlations of complementary skeleton motion, hand articulation and audio until the shared layer HLS.\vspace*{-2pt}

\subsubsection*{Initialization of the fusion process}
\label{sec:traininginitialization}

\begin{figure*}[!t]
%%\begin{subfigure}[b]{\linewidth}
\centering
\includegraphics[width = 130mm]{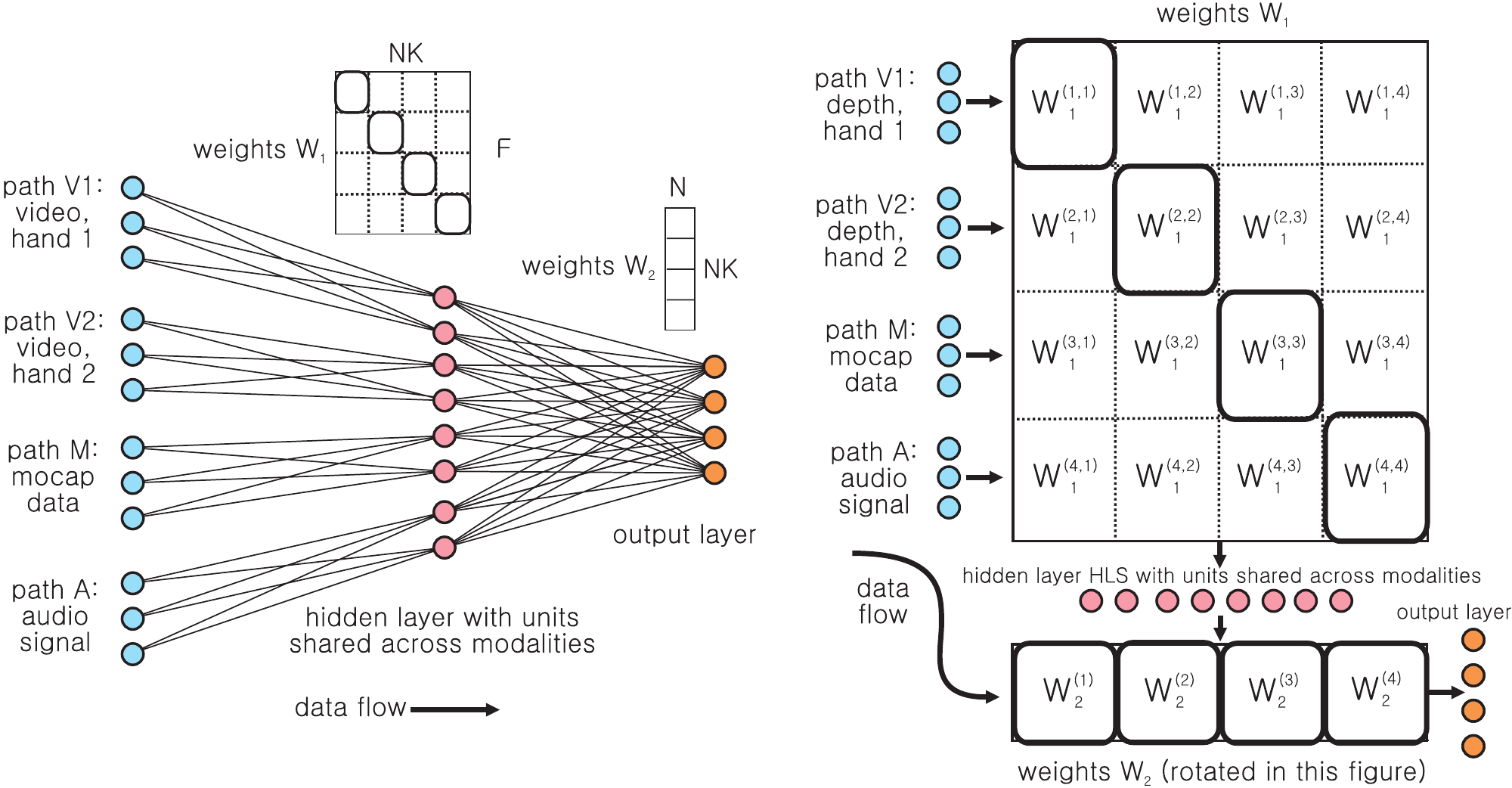}\vspace*{-8pt}
%	\label{fig:weights02}
%	\caption{}
%\end{subfigure}
   	\caption{On the left: architecture of shared hidden and output layers. %Output hidden units on each path are first connected to a subset of neurons of the shared layer. During fusion, additional connections between paths and the shared hidden layer are added (not shown).
On the right: structure of parameters of shared hidden and output layers (corresponds to the architecture on the left). 
%Note that the image scale is chosen for clarity of description and the real aspect ratio between dimensions of the matrix $W_1$ is not preserved (it has 1600 rows and 84 columns), the ratio between vertical sizes of matrix blocks corresponding to different modalities is 9:9:7:7. 
%On the right: energy structure of weights $W_1$ after
%raining. Diagonal blocks are dominated by individual modalities
%(right and left hands, articulated pose and audio) out of diagonal
%elements reflect cross-modality correlations.\gwt{Note that a) and b)
%  are referred to in the caption and the text, but are not shown in
 % the figure. The top and bottom figures are not separated enough,
 % it's hard to tell where one stops and the other starts (e.g. the
 % data flow arrow). It may be helpful to show dimensions of each layer
 % in the top figure, using the notation introduced in the text.}
 }
	\label{fig:weights0}\vspace*{-8pt}
\end{figure*}

Assuming the weights of the modality-specific paths are pre-trained, the next important issue is determining a fusion strategy. Pre-training solves some of the problems related to learning in deep networks with many parameters. However, direct fully-connected wiring of pre-trained paths to the shared layer in large-scale networks is not effective, as the high degrees of freedom afforded by the fusion process may lead to a quick degradation of pre-trained connections. We therefore proceed by initializing the shared layer such that a given hard-wired fusion strategy is performed, and then gradually relax it to more powerful fusion strategies.

A number of works have shown that among fusion strategies, the weighted arithmetic mean of per-model outputs is the least sensitive to errors of individual classifiers  \cite{alexandre2001}. It is often used in practice, outperforming more complex fusion algorithms. Considering the weighted mean as a simple baseline, we aim to initialize the fusion process with this starting point and proceed with gradient descent optimization towards an improved solution.
% the minimum in case if it is not reached.

Unfortunately, implementing the arithmetic mean in the case of early fusion and non-linear shared layers is not straightforward \cite{maxout}. %\gwt{Is there a reference for this?} \nn{There is a similar phrase in the Maxout paper with no further explanation (sounds like it is obvious) -- not sure whether it is worth citing.} \cw{Let's cite it.} 
It has been shown though \cite{BaldiSadowski20124}, that in dropout-like \cite{dropout} systems activation units of complete models produce a weighted normalized geometric mean of per-model outputs.  This kind of average approximates the arithmetic mean better than the geometric mean and the quality of this approximation depends on consistency in the neuron activation. We therefore initialize the fusion process to a normalized geometric mean of per-model outputs. 

Data fusion is implemented at two different layers: the shared hidden layer (HLS) and the output layer. The weight matrices of these two layers, denoted respectively as $W_1$ and $W_2$, are block-wise structured and initialized in a specific way, as illustrated in Fig.~\ref{fig:weights0}. The left figure shows the architecture in a conventional form as a diagram of connected neurons. The weights of the connections are indicated by matrices. %\gwt{It looks like in the figure, the weights are indicated by connected line segments, not matrices -- though there is a matrix and a vector above the line segments} 
%\nn{Do not understand. Line segments = connections, matrices above = weights, no?}
%\cw{It also looks clear to me...}
On the right we introduce a less conventional notation, which allows one to better visualize and interpret the block structure. Note that the image scale is chosen for clarity of description and the real aspect ratio between dimensions of $W_1$ ($1600{\times}84$) is not preserved, the ratio between vertical sizes of matrix blocks corresponding to different modalities is 9:9:7:7. 

We denote the number of hidden units in the modality- specific hidden layers on each path as $F_k$, where $k{=}1{\ldots}K$ and $K$ is the number of modality-specific paths. We set the number of units of the shared hidden layer equal to $K{\cdot}N$, where $N$ is the number of target gesture classes.%, and M is a hyper-parameter which governs model complexity.

As a consequence, the matrix $W_1$ of the shared hidden layer is of size $F{\times}(N{\cdot}K)$, where $F{=}\sum_k F_k$, and the weight matrix $W_2$ of the output layer is of size $(N{\cdot}K){\times}N$. Weight matrix $W_1$ can be thought of as a matrix of $K{\times}K$ blocks, where each block $k$ is of size $F_{k}{\times}N$. %\gwt{Ah, this notation has been greatly simplified and improved since the version earlier today! One comment though: in the previous paragraph, you have defined the \# of units in the shared layer as $M{\cdot}N{\cdot}K$, but $M$ does not show up in the dimensions of the weights here, and you have not said that you are setting $M=1$}  
%\nn{Christian, do you still want this M?}
%\cw{After some thought, we can probably drop it. Still think that we impose model complexity where it should be open, but it makes things complicated. Perhaps we could just say something on this.}
%\nn{It is difficult to explain this M given the logic of our initialization. If each block is simply repeated M times, we introduce symmetries in weights and it will not be useful. 
%So should be something more sophisticated... I don't think we want to go into it.}
This imposes a certain meaning on the units and weights of the network. Each column in a block (and each unit in the shared layer) is therefore related to a specific gesture class. Note that this block structure (and meaning) is forced on the weight matrix during initialization and in the early phases of training. If only the diagonal blocks are non-zero, which is forced at the beginning of the training procedure, then individual modalities are trained independently, and no cross correlations between modalities are captured. During the final phases of training, no structure is imposed and the weights can evolve freely. Formally, the activation of each hidden unit $h_{l}^{k}$ in the shared layer can be expressed as:\vspace*{-5pt}
\begin{equation} h_{l}^{(k)}\!\! = \!\sigma\!\left[\!\vphantom{\sum}\right. \sum_{i=1}^{F_k}\!w_{i,l}^{(k,k)}\!x_i^{(k)}\!\! + \!\gamma\!\! \sum_{\substack{m=1\\m\neq k}}^{K}\!\sum_{i=1}^{F_n}\!w_{i,l}^{(m,k)}x_i^{(m)}\!\! + b_{l}^{(k)}\! \left.\vphantom{\sum}\right]
\label{sec:defsharedhiddenlayer}
\end{equation} \vspace*{-10pt}\\
where $h_{l}^{(k)}$ is unit $l$ initially related to modality $k$, and all $w$ are from weight matrix $W_1$. Notation $w_{i,l}^{(m,k)}$ stands for a weight between non-shared hidden unit $i$ from the output layer of modality channel $m$ and the given shared hidden unit $l$ related to modality $k$. Accordingly, $x_i^{(m)}$ is input number $i$ from channel $m$, $\sigma$ is an activation function. Finally, $b_{l}^{(k)}$ is a bias of the shared hidden unit $h_l^{(k)}$.
The first term contains the diagonal blocks and the second term contains the off-diagonal weights. Setting $\gamma{=}0$ freezes learning of the off-diagonal weights responsible for inter-modality correlations.

%\begin{equation} %h_{j}^{(m)} = \sigma\left[ %\sum_{i=1}^{F_m}w_{i,j}^{(m)}x_i^{(m)} + %\sum_{n\neq m}\delta(X^{(n)}) + b_{j}^{(m)} %\right]
%\end{equation} %where $h_{j}^{(m)}$ is unit $j$ for modality $m$ $\delta(X(n)) = \sum_{i=1}^{F_n}w_{i,j}^{(n)}x_i^{(n)}$.

This initial meaning forced onto both weight matrices $W_1$ and $W_2$ produces a setting where the hidden layer is organized into $K$ subsets of units $h_{l}^{(k)}$, one for each modality $k$, and where each subset comprises $N$ units, one for each gesture class. The weight matrix $W_2$ is initialized in a way such that these units are interpreted as posterior probabilities for gesture classes, which are averaged over modalities by the output layer controlled by weight matrix $W_2$. In particular, each of the $N{\times}N$ blocks of the matrix $W_2$ (denoted as $v\symbk$) is initialized as an identity matrix, which results in the following expression for the output units, which are softmax activated:\vspace*{-3pt}
\begin{equation} o_j = \dfrac {e^{\sum_{k=1}^K \sum_{c=1}^N v\symbk_{j,c} h\symbk_c}} {\sum_{i=1}^N e^{\sum_{k=1}^K \sum_{c=1}^N v\symbk_{i,c} h\symbk_c}} = \dfrac {e^{\sum_{k=1}^K h\symbk_j}} {\sum_{i=1}^N e^{\sum_{k=1}^K h\symbk_i}}
\label{eq:w2doesgeomean}
\end{equation} \vspace*{-8pt}\\
where we used that $v_{j,c}\symbk{=}1/K \textrm{ if } j{=}c$ and $0$ else.

From (\ref{eq:w2doesgeomean}) we can see that the diagonal initialization of $W_2$ forces the output layer to perform modality fusion as a normalized geometric mean over modalities, as motivated in the initial part of this section. Again, this setting is forced in the early stages of training and relaxed later, freeing the output layer to more complex fusion strategies.

%\cw{Weights $w$ here have the same symbol for $W_1$ and $W_2$. Not good, need to discriminate}.

\subsubsection*{ModDrop: multimodal dropout}

\noindent Inspired by the concept of dropout \cite{dropout} as the normalized geometric mean of an exponential number of weakly trained models, we aim on exploiting \emph{a priori} information about groupings in the feature set. We initiate a similar process but with a fixed number of models corresponding to separate modalities and pre-trained to convergence. We have two main motivations: (i) to learn a shared model while preserving uniqueness of per-channel features and avoiding false co-adaptations between modalities; (ii) to handle missing data in one or more of the channels at test time. The key idea is to train the shared model in a way that it would be capable of producing meaningful predictions from an arbitrary number of available modalities (with an expected loss in precision when some signals are missing).

Formally, let us consider a set of $\mathcal{M}\symbk$, $k{=}1{\ldots}K$ modality-specific models. During pretraining, the joint learning objective can be generally formulated as follows:\vspace*{-5pt}
\begin{equation}
\label{eq:pretrainingloss}
\mathcal{L}_{\mathrm{pretraining}} = \sum_{k=1}^K \mathcal{L}\left[\mathcal{M}^{(k)}\right] + \alpha\sum_{h=1}^{H}||W_h||^2,
\end{equation}\vspace*{-10pt}\\
\noindent where each term in the first sum represents a loss of the corresponding modality-specific model (in our case, negative log likelihood, summarized over all samples $x_d$ for the given modality $k$ from the training set $|\mathcal{D}|$):\vspace*{-5pt}
\begin{equation}
\mathcal{L}\left[\mathcal{M}^{(k)}\right] = -\sum_{d\in\mathcal{D}}\log{o^{(k)}_{Y}(Y=y_d|x_d^{(k)})},
\end{equation}\vspace*{-10pt}\\
where $o^{(k)}_{Y}$ is output probability distribution over classes of the network corresponding to modality $k$ and $y_d$ is a ground truth label for a given sample $d$.
%\cw{If the distribution is not just called p (like "probability"), then I think we should say a couple of words about it, so, what is $o_Y$ a part from plain log likelihood? Also: $o_j$ also denotes output $j$.}

The second term in Eq.~\ref{eq:pretrainingloss} is $L_2$ regularization on all weights $W_h$ from all hidden layers $h{=}1{\ldots}H$ in the network (with weight $\alpha$).
At this pretraining stage, all loss terms in the first sum are minimized independently.
% for each data channel.

Once the weight matrices $W_1$ and $W_2$ are initialized with pre-trained diagonal elements and initially zeroed out off-diagonal blocks of weights are relaxed (i.e.~$\gamma{=}1$ in Eq.~\ref{sec:defsharedhiddenlayer}), fusion is learned from the training data. The desired training objective during the fusion process can be formulated as a combination of losses of all possible combinations of modality-specific models:\vspace*{-5pt}
\begin{align}
\label{eq:combinedloss}
\!\!\!\mathcal{L}_{\Sigma} & \!=\!\!
\sum_{k=1}^K\!\mathcal{L}\!\left[\!\mathcal{M}^{(k)}\!\right]\! + \!\!\!\sum_{k\neq m}\!\! \mathcal{L}\!\left[\!\mathcal{M}^{(k,m)}\!\right] %\nonumber \\
 %& 
 \!+\!\!\!\!\!\! \sum_{k\neq m \neq n}\!\!\!\!\! \mathcal{L}\!\left[\!\mathcal{M}^{(k,m,n)}\!\right]\! +\! {\ldots} \nonumber\vspace*{-35pt}\\
& + \alpha\sum_{h=1}^H\!||W_h||^2\!=\!
\sum_{S\in\mathcal{P}({\mathcal{M}}^{(k)})}^{} \!\!\!\mathcal{L}\left[S\right] + \alpha\sum_{h=1}^{H}||W_h||^2 ,
\end{align}\vspace*{-10pt}\\
\noindent where  $\mathcal{M}^{(k,m)}$ indicates fusion of models $\mathcal{M}^{(k)}$, $\mathcal{P}(\mathcal{M}^{(k)})$ is the powerset of all models, whose size is $2^{K}$, and $S$ is an element of the power set  corresponding to all possible combinations of modalities. 

The loss function formulated in (\ref{eq:combinedloss}) reflects the objective of the training procedure, but in practice we approximate this objective by ModDrop as iterative interchangeable training of one term at a time. In particular, the fusion process starts by joint training through backpropagation over the shared layers and fine tuning all modality specific paths.
%Each modality pathway $m$ is dropped with the probability $p_m$ (in our experiments set to $0.25$). %\gwt{Is it necessary to introduce notation for separate dropout probabilities per-modality? It is never used} 
As this step, the network takes as an input multi-modal training samples $\{\delta^{(k)}x_d^{(k)}\}$, $k=1{\ldots}K$ from the training set $|\mathcal{D}|$, where for each sample each modality component $x_d^{(k)}$ is dropped (set to $0$) with a certain probability $q\symbk{=}1{-}p\symbk$ %where $p\symbk$ is 
indicated by Bernoulli selector 
%\begin{equation}
$\delta^{(k)}{:}\,P(\delta^{(k)}{=}1){=}p^{(k)}$.
%\end{equation}
Accordingly, one step of gradient descent given an input with a certain number of non-zero modality components minimizes the loss of a corresponding multi-modal subnetwork denoted as $\{\delta^{(k)}\mathcal{M}^{(k)}\}$.
This aligns well with the initialization process described above, which ensures that modality-specific subnetworks that are being removed or added by ModDrop are well pre-trained in advance.

\subsubsection*{Regularization properties}

\begin{figure}[!t]
	\begin{center}
		\includegraphics[width=70mm]{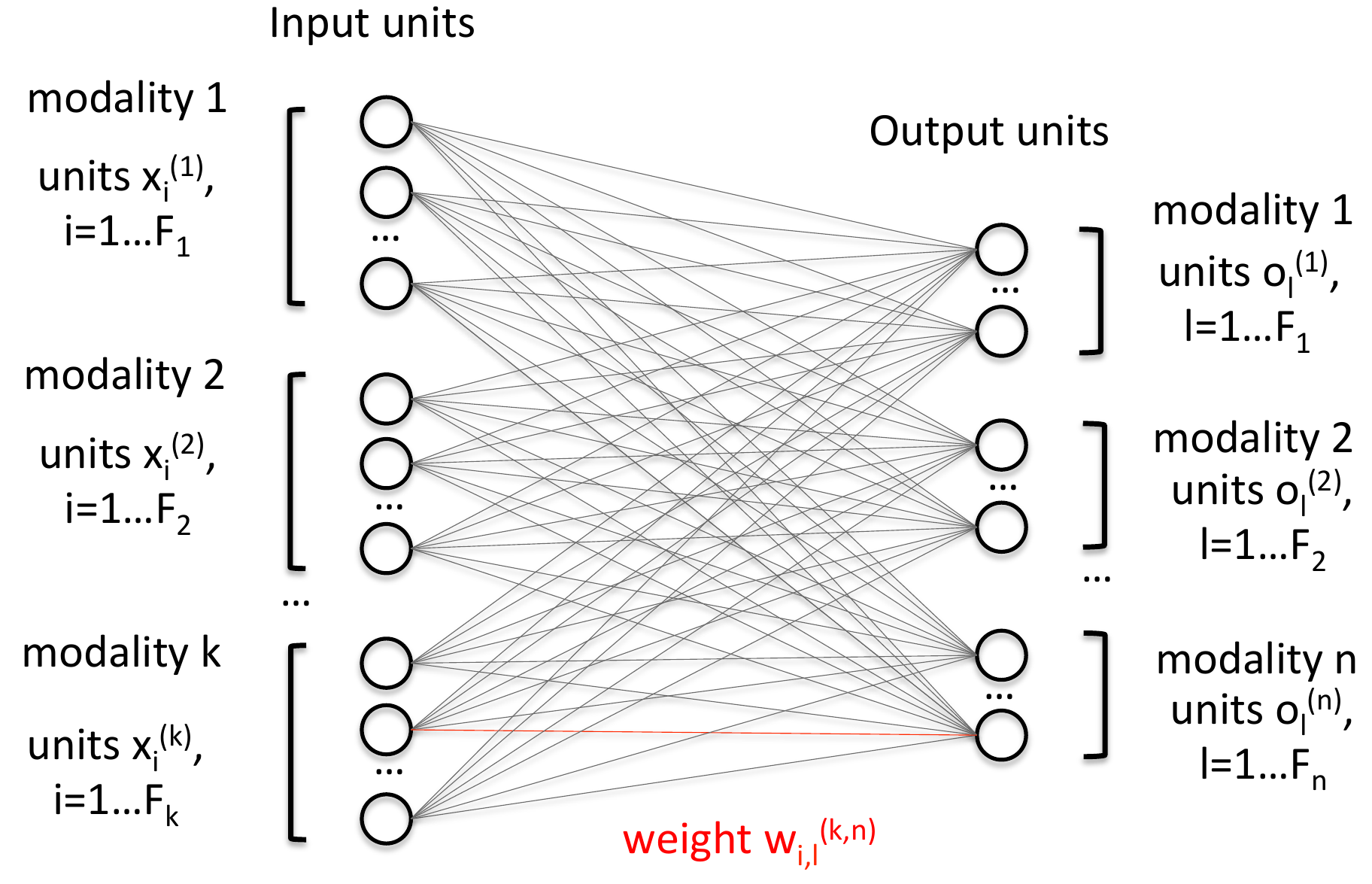}
	\end{center}\vspace*{-8pt}
  	\caption{Toy network architecture and notations used for derivation of ModDrop regularization properties.}
	\label{fig:toynotations}\vspace*{-12pt}
\end{figure}

\noindent 
%\nn{Start reading here.}
%\text{Original attempt}
In the following we will study the regularization properties of
modality-wise dropout on inputs (ModDrop) on a simpler network architecture, namely a
one-layer shared network with $K$ modality specific paths and sigmoid
activation units. Input $i$ for modality $k$ is denoted as $x_i\symbk$
and we assume that there are $F_k$ inputs coming from each
modality~$k$ (see Fig. \ref{fig:toynotations}). Output unit $l$ related to modality $n$ is denoted as $o_{l}\symbn$. Finally, a weight coefficient connecting input unit $x_{i}\symbk$ with output unit $o_{l}\symbn$ is denoted as $w_{i,l}^{(k,n)}$.

In our example, output units are sigmoidal, i.e. for each output unit $o_l$ related to modality $n$, $o_l\symbn = \sigma(s_{l}\symbn) =
1/(1+e^{-\lambda {s_{l}\symbn}})$, where $s_l\symbn{=}\sum_{k=1}^{K}\sum_{i}^{F_k}w_{i,l}^{(k,n)}x_i^{(k)}$ is the input to the activation
function coming to the given output unit from the previous layer, and $\lambda$ is a coefficient. %\gwt{I'm not sure I like this
%  mixed notation of lower case and upper case for scalars, e.g. $I, S,
 % h, y$. Apart from error $E$, I don't think we've used upper case for
 % scalars up to this point. Also: I think it needs to be made explicit at some point here you are analyzing a single unit.}

We minimize cross-entropy error calculated from the targets $y$ (indices are dropped for simplicity)\vspace*{-5pt}
\begin{equation}
E=-(y\log{o} + (1-y)\log{(1-o)}),
\end{equation}\vspace*{-18pt}\\
whose partial derivatives can be given as follows:
\begin{equation*}
\dfrac{\partial E}{\partial w} = \dfrac{\partial E}{\partial o} \dfrac{\partial o}{\partial s} \dfrac{\partial s}{\partial w},\quad
%\end{equation}
%where
%\begin{equation*}
\dfrac{\partial E}{\partial o} = -y\dfrac{1}{o} + (1-o)\dfrac{1}{1-o},
\end{equation*}\vspace*{-10pt}
\begin{equation}
\dfrac{\partial o}{\partial s} = \lambda o (1-o),\quad
\dfrac{\partial E}{\partial w} = -\lambda (y-o)\dfrac{\partial s}{\partial w}.
\end{equation}\vspace*{-10pt}\\
%\gwt{Another notational thing that could confuse neural nets people
 % here, is that you're calling the units `Output units', and defining
 % Error on them like outputs, but using the
 % symbol $h$ which is typically used for hidden units} 
Along the lines of \cite{BaldiSadowski20124}, we consider two situations corresponding to two different loss functions: $E_{\Sigma}$, corresponding to the ``complete network'' where all modalities are present, and $\tilde{E}$ where ModDrop is performed. In our case, we assume that whole modalities (sets of units corresponding to a given modality $k$) are either dropped %(with probability $q\symbk=1-p\symbk$) 
or preserved. 
%(with probability $p\symbk$). 
In a ModDrop network, this can be formulated such that the input to the activation function of a given output unit $l$ related to modality $n$ (denoted as $\tilde{s}_{l}\symbn$) involves a Bernoulli selector variable $\delta\symbk$ for each modality $k$ which can take on values in $\{0,1\}$ and is activated with probablity $p\symbk$:\vspace*{-5pt}
\begin{equation}
\tilde{s}_{l}\symbn=\sum_{k=1}^K\delta\symbk\sum_{i=1}^{F_k} w_{i,l}^{(k,n)}x_i\symbk
\end{equation}\vspace*{-10pt}\\
As a reminder, in the case of the complete network (all channels are present) the output activation it the following:% \gwt{You need to explain what you mean by `complete network'}:
\vspace*{-5pt}\begin{equation}
s_{l}\symbn=\sum_{k=1}^K\sum_{i=1}^{F_k} w_{i,l}^{(k,n)}x_i^{(k)}
\end{equation}\vspace*{-20pt}\\
\noindent
As the following reasoning always concerns a single output unit $l$ related to modality $n$, from now on these indices will be dropped for simplicity of notation. Therefore, we denote $s=s_{l}\symbn$,  $\tilde{s}=\tilde{s}_{l}\symbn$ and $w_{i}\symbk=w_{i,l}^{(k,n)}$.

Gradients of corresponding complete and ModDrop sums with respect to weights can be expressed as follows:\vspace*{-5pt}
\begin{equation}
\dfrac{\partial \tilde{s}}{\partial w_i\symbk} = \delta\symbk x_i\symbk, \quad
\dfrac{\partial s}{\partial w_i\symbk} = x_i\symbk %\quad
\end{equation}\vspace*{-10pt}\\
Using the gradient of the error $E$ \vspace*{-5pt}
\begin{equation}
\dfrac{\partial E}{\partial w_i\symbk} = 
-\lambda\left[y-\sigma(s)\right] 
\dfrac{\partial s}{\partial w_i\symbk},
\end{equation}\vspace*{-10pt}\\
the gradient of the error for the complete network is:\vspace*{-5pt}
\begin{equation}
\label{eq:gradientcompletenetwork}
\dfrac{\partial E_{\Sigma}}{\partial w_i\symbk} = 
-\lambda x_i\symbk\!\left[y\vphantom{\sum}\right.-\sigma{\left(\vphantom{\sum}\right.\sum_{m=1}^K\sum_{j=1}^{F_m} w_j\symbm x_j\symbm\left.\!\vphantom{\sum}\right)}\left.\!\!\vphantom{\sum}\right]
\end{equation}\vspace*{-10pt}\\
In the case of ModDrop, for one realization of the network where a modality is dropped with corresponding probability $q\symbk{=}1{-}p\symbk$, indicated by the means of Bernoulli selectors $\delta\symbk$, i.e. $P(\delta\symbk{=}1){=}p\symbk$, we get:\vspace*{-5pt}
\begin{equation}
\!\dfrac{\partial \tilde{E}}{\partial w_i\symbk}\! =\!
-\lambda\delta\symbk\!x_i\symbk\!\left[\vphantom{\sum}\right.\!y-\sigma{\left(\!\vphantom{\sum}\right.\sum_{m=1}^K\!\!\delta\symbm\!\sum_{j=1}^{F_m} w_j\symbm \!x_j\symbm\!\left.\vphantom{\sum}\right)}\!\!\left.\vphantom{\sum}\right]\!\!\!\!
\end{equation}\vspace*{-10pt}\\
Taking the expectation of this expression requires an expression introduced in \cite{BaldiSadowski20124}, which approximates $E[\sigma(x)]$ by $\sigma(E[x])$. We take the expectation over the $\delta\symbm$ with the exception of $\delta\symbk{=}1$, which is the Bernouilli selector of the modality $k$ for which the derivative is calculated:\vspace*{-5pt}
\begin{align*}
E\left[\!\vphantom{\tilde{\sum}}\right.\dfrac{\partial \tilde{E}}{\partial w_i\symbk}\left.\vphantom{\tilde{\sum}}\!\!\right] & 
\!\approx\! -\lambda p\symbk\! x_i\symbk\!\left[y\vphantom{\sum}\right.-\sigma\!\left(\!\vphantom{\sum}\right.\sum_{m\neq k}^{K} p\symbm\!\sum_{j=1}^{F_m} w_j\symbm\! x_j\symbm\nonumber\\
&\hspace*{-52pt}+\!\sum_{j=1}^{F_k}\! w_j\symbk\! x_j\symbk 
\!\left.\vphantom{\sum}\right)\!\left.\vphantom{\sum}\right]%\nonumber\\
%& \hspace{-50pt}
\!=\! -\lambda p\symbk x_i\symbk\left[y\vphantom{\sum}\right.\!-\!\sigma\!\left(\vphantom{\sum}\!\right.\sum_{m\neq k}^{K}\sum_{j=1}^{F_m} w_j\symbm \!x_j\symbm\nonumber\\
&\hspace*{-52pt}- \!\!\sum_{m\neq k}^{K}\!(1\!-\!p\symbm)\!\sum_{j=1}^{F_m} w_j\symbm\! x_j\symbm\!\! +\!\!
\sum_{j=1}^{F_k} w_j\symbk \!x_j\symbk 
\!\!\left.\vphantom{\sum}\right)\!\!\left.\vphantom{\sum}\right]\!\! =\!-\lambda p\symbk\! x_i\symbk \nonumber\\
%\end{align*}
%\begin{align*}
%E\left[\dfrac{\partial E_{md}}{\partial w_i\symbk}\right] & 
&\hspace{-52pt}%=\!-\lambda\!\left[y\!
\times\!\!\left[y\vphantom{\sum}\right.\!-\!\sigma\!\left(\!\vphantom{\sum}\right.\!\sum_{m=1}^{K}\!\sum_{j=1}^{F_m} w_j\symbm\!x_j\symbm %-\right.\right.\nonumber\\
%&\hspace*{-57pt} \left.\left. 
\!\!\!-\!\!\sum_{m\neq k}^{K} \!(1\!-\!p\symbm)\!\sum_{j=1}^{F_m} w_j\symbm\!x_j\symbm  
\!\left.\vphantom{\sum}\right)\!\!\left.\vphantom{\sum}\right]
%\\
%&\hspace*{-20pt}=
%-\lambda\left[y-\sigma\left(\sum_{m=1}^{M} p\symbm \sum_{j} w_j\symbm x_j\symbm \right.\right.\\
%&\hspace*{-55pt}\left.\left. + (1-p\symbk)\!\sum_{j\neq i}w_j\symbk x_j\symbk\! +
%(1-p\symbk)w_i \symbk x_i \symbk\right)\right]p\symbk x_i \symbk
\end{align*}

%\right)}\right]I_i^{(m)} $$
Taking the first-order Taylor expansion of the activation function $\sigma$ around $s = \sum_{m} \sum_{j} w_j\symbm\!x_j\symbm$ gives\vspace*{-5pt}
\begin{align*}
\centering
\!E\!\!\left[\!\vphantom{\tilde{\sum}}\right.\!\dfrac{\partial \tilde{E}}{\partial w_i\symbk}\!\left.\!\vphantom{\tilde{\sum}}\!\right]\!\! &\approx\!\!-\lambda p\symbk\!x_i\symbk \!\!\left[\!\vphantom{\sum}\right.%\vphantom{\sum_{j\neq i}w_j^{(m)}x_j^{(m)}}
%\vphantom{\sum_{j\neq i}w_j^{(m)}x_j^{(m)}}
\!y\!-\!\sigma_s\!%\right.%\nonumber\\
%&\hspace*{-55pt}
%\left.\!
\!+\!\sigma_s'\!%
\!\sum_{m\neq k}^{K}\!\!(1\!-\!p\symbm\!)\!\!\sum_{j=1}^{F_m}\!\!w_j\symbm\!x_j\symbm
%+(1-p\symbk)w_i\symbk x_i\symbk%
\left.\vphantom{\sum}\!\!\right]
\end{align*}
where $\sigma'_s\!{=}\sigma'(s)\!{=}\sigma(s)\!{/}\!(1{-}\sigma(s))$.
Substituting equation (\ref{eq:gradientcompletenetwork}),\vspace*{-5pt}
\begin{align*}
\!\!E\!\!\left[\!\vphantom{\tilde{\sum}}\right.\!\dfrac{\partial \tilde{E}}{\partial w_i\symbk}\!\left.\vphantom{\tilde{\sum}}\!\!\right]\!\!\approx\! p\symbk\!\!\dfrac{\partial E_{\Sigma}}{\partial w_i\symbk} %\nonumber\\
%&\hspace*{-40pt}
\!-\!\lambda\sigma_{\!\!s}'\!x_i\symbk \!p\symbk\!\!\!\sum_{m\neq k}^{K}\!\!(1\!-\!p\symbm\!)\!\!\sum_{j=1}^{F_k}\!w_j \symbm\! x_j\symbm
\end{align*}
If $p\symbk{=}p\symbm{=}p$ then $p(1{-}p){=}\textrm{Var}(\delta)$. From the gradient, we can calculate the error $\tilde{E}$ integrating out the partial derivatives and summing over the weights $i$:\vspace*{-2pt}
\begin{align}
\label{eq:regresult}
\!\!\tilde{E}\!\approx\!pE_{\Sigma}\!
-\!\! \lambda\sigma_{\!\!s}'\textrm{Var}(\delta)\!\!
\sum_{k=1}^{K}\!\sum_{m\neq k}^{K}\!\sum_{i{=}1}^{F_k}\!\sum_{j=1}^{F_m}\!\!
w_i\symbk\! w_j \symbm\! x_i \symbk\! x_j\symbm
\end{align}\vspace*{-8pt}\\
As it can be seen,
% from the final expression ($\ref{eq:regresult}$), 
the error of the network with ModDrop is approximately equal to the error of the complete model (up to a coefficient) minus an additional term including a sum of products of inputs and weights corresponding to different modalities in all possible combinations. We need to stress here that this second term reflects exclusively cross-modality correlations and does not involve multiplications of inputs from the same channel.
To understand what influence the cross-product term has on the training process, we analyse two extreme cases depending on whether or not signals in different channels are correlated.

Let us consider two input units $x_i\symbk$ and $x_j\symbm$ coming
from different modalities and first assume that they are independent and therefore uncorrelated. Since each network input is normalized to zero mean, the expectation is also equal to zero:\vspace*{-5pt}
\begin{equation}
E\!\left[\!\!\vphantom{\tilde{x}}\,\right.x_i\symbk\!x_{j}\symbm\left.\vphantom{\tilde{x}}\!\!\right]{=}E\!\left[\!\vphantom{\tilde{x}}\right.x_i\symbk\left.\vphantom{\tilde{x}}\!\!\right]\!E\!\left[\vphantom{\tilde{x}}\,\right.\!\!x_j\symbm\left.\vphantom{\tilde{x}}\!\!\right]{=}0.
\end{equation}\vspace*{-15pt}\\
Weights in a single layer of a neural network typically obey a unimodal distribution with zero  expectation \cite{fastdropout2013}. It can be shown~\cite{lehmann1998} that under these assumptions, Lyapunov's condition is satisfied and that Lyapunov's central mean theorem holds; in this case the sum of products of inputs and weights will tend to a normal distribution given that the number of training samples is sufficiently large. As both the input and weight distributions have zero mean, the resulting law is also centralized and its variance is defined by the magnitudes of the weights (assuming inputs are fixed).

We conclude that, assuming independence of inputs in different channels, the second term in equation ($\ref{eq:regresult}$) tends to vanish if the number of training samples in a batch is sufficiently large. In practice, additional regularization on weights is required to prevent weights from exploding.
%The difference with the result from the original attempt is that 
%1) we don't have dropout like L2 regularization (which is good as it justifies why ModDrop does not work without dropout, or potentially any other way of weight regularization);
%2) the cross-weight "regularizations" are not inside modalities as before (and not inside out of diagonal blocks as Christian proposed), but between blocks belonging to different modalities. I.e. this time it really reflects cross-modality relations.

%\nn{Message to take from the paragraph below is that the second term in the expression above obeys normal distribution with zero mean and variance defined by magnitudes of weights. I.e. for large training batches its expectation is 0, for a single example can easily diverge. Additional regularization on weights is required to control it.}

%\nn{Let's consider two extreme cases for each pair of input units  belonging to different modalities and therefore participating in the product.}

Now let us consider a more interesting scenario when two inputs $x_i\symbk$ and $x_j\symbm$  belonging to different modalities are positively correlated. In this case, given zero mean distributions on each input, their product is expected to be positive:\vspace*{-5pt}
\begin{equation}
E\!\left[\!\!\vphantom{\tilde{x}}\,\right.x_i\symbk\!x_{j}\symbm\left.\vphantom{\tilde{x}}\!\!\right]{=}E\!\left[\!\vphantom{\tilde{x}}\right.x_i\symbk\left.\vphantom{\tilde{x}}\!\!\right]\!E\!\left[\vphantom{\tilde{x}}\,\right.\!\!x_j\symbm\left.\vphantom{\tilde{x}}\!\!\right] + \text{Cov}\!\left[\!\vphantom{\tilde{x}}\right.x_i\symbk\!\!,x_j\symbk\left.\vphantom{\tilde{x}}\!\right]\!.
\end{equation}\vspace*{-15pt}\\
Therefore, on each step of gradient descent this term enforces the product $w_i\symbk w_j\symbm$ to be positive and therefore introduces correlations between these weights (given, again, the additional regularization term preventing one of the multipliers from growing significantly faster than the other). The same logic applies if inputs are negatively correlated, which would enforce negative correlations on corresponding weights. Accordingly, for correlated modalities this additional term in the error function introduced by ModDrop acts as a cross-modality regularizer forcing the network to generalize by discovering similarities between different signals and ``aligning'' them with each other by introducing soft ties on the corresponding weights.

Finally, as has been shown by \cite{BaldiSadowski20124} for dropout, the multiplier proportional to the derivative of the sigmoid activation makes the regularization effect adaptive to the magnitude of the weights. As a result, it is strong in the mid-range of weights, plays a less significant role when weights are small and gradually weakens with saturation. 

Our experiments have shown that ModDrop achieves the best results if combined with dropout, which introduces an adaptive L2 regularization term $\hat{E}$ in the error function~\cite{BaldiSadowski20124}:\vspace*{-5pt}
\begin{equation}
\hat{E} \approx \lambda\sigma_{s}'\text{Var}(\hat{\delta})\sum_{k=1}^{K}\sum_{i=1}^{F_k}\left[w_i\symbk x_i\symbk\right]^2\!\!\!, 
\end{equation}\vspace*{-10pt}\\
where $\hat{\delta}$ is a Bernoulli selector variable, $P(\hat{\delta}{=}1){=}\hat{p}$ and $\hat{p}$ is the probability that a given input unit is present.

\section{Inter-scale fusion during test time}
\label{fusion}
%Different models, late fusion by summarizing predictions. %Ok, convolutions. Contribution of each frame to the final prediction at the moment $t_0$ (from neural models):
%\begin{align}
 %f(t')=\sum_{t=1}^{T}\textrm{rect}\,(t-t_0)\sum_{s=2}^{4}\mu_s\textrm{rect}\,\left(\dfrac{t'-t}{4s+1}\right) = \sum_{s=2}^4 \mu_s(4s+1)\Lambda\,\left(\dfrac{t'-t_0}{4s+1}\right)
%\end{align}

Once individual single-scale predictions are obtained, we employ a simple voting strategy for fusion with a single weight per model. We note here that introducing additional per-class per-model weights and training meta-classifiers (such as an MLP) on this step quickly leads to overfitting.

At each given frame $t$, per-class network outputs $o_k$ are obtained via per-frame aggregation and temporal filtering of predictions at each scale with corresponding weights $\mu_s$ defined empirically: \vspace*{-9pt}%through cross-validation:
\begin{align}
o_k(t) =  \sum_{s=2}^4 \mu_s\sum_{j=-4s}^0 o_{s,k}(t+j),
\end{align}\vspace*{-10pt}\\
where $o_{s,k}(t+j)$ is the score of class $k$ obtained for a spatio-temporal block sampled starting from the frame $t+j$ at step $s$.
Finally, the frame is assigned the class label $l(t)$ having the maximum score:
%\begin{align}
$l(t) = \argmax_k{o_k(t)}$.
%\end{align}
%Per-frame prediction aggregation. More sophisticated fusion (an MLP, random search) -- quick overfitting.

%%% Local Variables: 
%%% mode: latex
%%% TeX-master: "main"
%%% End: 

%\begin{figure}[!t]
%\centering
%\includegraphics[height=0.6\linewidth]{localization_.eps}
%\caption{Gesture localization. Top: output predictions of the main classifier; Bottom: output of the binary motion detector. Noise at pre-stroke and post-stroke phases in the first case is due to high similarity between gesture classes at these time periods and temporal inertia of the classifier.}
%\label{fig:localization}
%\end{figure}

\section{Gesture localization}
\label{sec:localization}

With increasing duration of a dynamic pose, recognition rates of the classifier increase at a cost of loss in precision in gesture localization. Using wider sliding windows leads to noisy predictions at pre-stroke and post-stroke phases due to the overlap of several gesture instances at once.
On the other hand, too short dynamic poses are not discriminative either, as most gesture classes at their initial and final stages have a similar appearance (e.g.~raising or lowering hands). 

There exists vast literature on temporal video segmentation (\cite{fang2007}), however in this work we employ a simpler and yet efficient solution. To address this issue, we introduce an additional binary classifier to distinguish resting moments from periods of activity. Trained on dynamic poses at the finest temporal resolution $s{=}1$, this classifier is able to precisely localize starting and ending points of each gesture.

The module is a two-layer fully connected network taking as an input the articulated pose descriptor.
All training frames having a gesture label are used as positive examples, while 
a set of frames right before and after such gesture are considered as negatives. Each frame is thus assigned with a label ``motion" or ``no motion" with accuracy of $98\%$.
 
To combine the classification and localization modules, frame-wise gesture class predictions are first obtained as described in Section \ref{fusion}. Output predictions at the beginning and at the end of each gesture are typically noisy. %(illustrated by the top curve at Fig.~\ref{fig:localization}). 
Therefore, for each spotted gesture, its boundaries are extended or shrunk towards the closest switching point produced by the binary classifier (assuming that this point is in a vicinity of the initially detected boundary). 

\section{Experiments}
\label{sec:experiments_}

The \textit{Chalearn 2014 Looking at People Challenge (track 3)} dataset  \cite{escalera2014chalearn} consists of 13,858 instances of Italian conversational gestures performed by different people and recorded with a consumer RGB-D sensor. It includes color, depth video and mocap streams. The gestures are drawn from a large vocabulary, from which 20 categories are identified to detect and recognize and the rest are considered as arbitrary movements. 
Each gesture in the training set is accompanied by a ground truth label as well as information about its start- and end-points.
For the challenge, the corpus was split into development, validation and test sets. The test data was released to participants after submitting their source code.

To further explore the dynamics of learning in multi-modal systems, we augmented the data with audio recordings extracted from a dataset released under the framework of the \textit{Chalearn 2013 Multi-modal Challenge on Gesture Recognition} \cite{escalera2013chalearn}. Differences between the 2014 and 2013 versions are mainly permutations in sequence ordering, improved quality of gesture annotations, and a different metric used for evaluation: the Jaccard index in 2014 instead of the Levenshtein distance in 2013.
As a result, each gesture in a video sequence is accompanied by a corresponding vocal phrase bearing the same meaning. Due to dialectical and personal differences in pronunciation and vocabulary, gesture recognition from the audio channel alone was surprisingly challenging. 

To summarize, we report results for two settings: %\vspace*{-3pt}
%\begin{itemize}
%\item 
i) the original dataset used for the \emph{ChaLearn 2014 Looking at People (LAP) Challenge} (track 3),
ii) an extended version of the dataset augmented with audio recordings taken from the \emph{Chalearn 2013 Multi-modal Gesture Recognition} dataset. \vspace*{-5pt}%\nn{link}
%\end{itemize}

\subsection{Experimental setup}

%\setlength{\tabcolsep}{4pt}
%\begin{table}[!t]
%\begin{center}
%\caption{\label{table:weights}
%Model weighing
%}
%\begin{tabular}{c|c}
%\hline\noalign{\smallskip}
%Model & Weight $\mu$ \\
%\noalign{\smallskip}
%\hline
%\noalign{\smallskip}
%Neural, step 2 & $0.26$ \\
%Neural, step 3 & $1.02$\\
%Neural, step 4 & $2.20$\\
%Extremely randomized trees, step 4 & $1.00$\\
%\hline
%\end{tabular}
%
%\end{center}
%\end{table}
%\setlength{\tabcolsep}{1.4pt}

Hyper-parameters of the multi-modal neural network for classification are provided in Table \ref{table:networkparameters}. The architecture is identical for each temporal scale. Gesture localization is performed with another MLP with 300 hidden units (see Section \ref{sec:localization}). All hidden units in the classification and localization modules have hyperbolic tangent activations. Hyper-parameters were optimized on the validation data with early stopping to prevent the models from overfitting and without additional regularization. 
For simplicity, fusion weights for the different temporal scales are set to $\mu_s{=}1$, as well as the weight of the baseline model (see Section \ref{fusion}). The deep learning architecture is implemented with the Theano library \cite{Theano}. A single-scale predictor operates at frame rates close to real time (24 fps on GPU).%\medskip

%\setlength{\tabcolsep}{4pt}
%\begin{table}[!t] \centering
%\caption{\label{table:networkparameters} Hyperparameters chosen for the deep learning models.}
%\vspace*{-2mm}
%\begin{tabular}{c|c|c|c}
%\hline\noalign{\smallskip}
%Layer & Filter / image size & Number of units & Pooling \\
%\noalign{\smallskip}
%\hline
%\multicolumn{4}{c}{Paths V1, V2} \\
%\hline
%Input D1,D2  & $72{\times}72{\times}5$ & & $2{\times}2{\times}1$\\
%ConvD1 & $25{\times}5{\times}5{\times}3$ & & $2{\times}2{\times}3$\\
%ConvD2 & $25{\times}5{\times}5$ & & $1{\times}1$\\
%Input C1,C2  & $72{\times}72{\times}5$ & &  $2{\times}2{\times}1$\\
%ConvC1 & $25{\times}5{\times}5{\times}3$ & & $2{\times}2{\times}3$\\
%ConvC2 & $25{\times}5{\times}5$ & & $1{\times}1$\\
%HLV1     & & $900$ &\\
%HLV2    & & $450$ &\\
%\hline
%\multicolumn{4}{c}{Path M} \\
%\hline
%Input M & & $183$ &\\
%HLM1    & & $700$ &\\
%HLM2   & & $700$ &\\
%HLM3   & & $350$ &\\
%\hline
%\multicolumn{4}{c}{Path A}\\
%\hline
%Input A  & $40{\times}9$ & & $1{\times1}$\\
%ConvA1 & $25{\times}5{\times}5{\times}3$ & & $1{\times}1$\\
%HLA1  & & $700$ &\\
%HLA2  & & $350$ &\\
%\hline
%\multicolumn{4}{c}{Shared layers}\\
%\hline
%HLS1 & & $1600$ &\\
%HLS2 & & $84$ &\\
%Output layer& & $21$ &\\
% \hline
%\end{tabular}
%\end{table}
%\setlength{\tabcolsep}{1.4pt}

\setlength{\tabcolsep}{4pt}
\begin{table}[!t] \centering
\begin{tabular}{c|c|c|c|c}
%\hline%\noalign{\smallskip}
Path & Layer & Filter size / \# units & \# parameters & Pooling \\
%\noalign{\smallskip}
\hline
%\multicolumn{4}{c}{Paths V1, V2} \\
%\hline
\multirow{8}{*}{\rotatebox{90}{Paths V1, V2}}&Input D1,D2  & $72{\times}72{\times}5$ & - & $2{\times}2{\times}1$\\
&ConvD1 & $25{\times}5{\times}5{\times}3$ & 1900 & $2{\times}2{\times}3$\\
&ConvD2 & $25{\times}5{\times}5$ & 650 & $1{\times}1$\\
&Input C1,C2  & $72{\times}72{\times}5$ & - &  $2{\times}2{\times}1$\\
&ConvC1 & $25{\times}5{\times}5{\times}3$ & 1900& $2{\times}2{\times}3$\\
&ConvC2 & $25{\times}5{\times}5$ & 650 & $1{\times}1$\\
&HLV1     & $900$& 3 240 900 & -\\
&HLV2    & $450$ & 405 450 & - \\
\hline
%&\multicolumn{4}{c}{Path M} \\
%\hline
\multirow{4}{*}{\rotatebox{90}{Path M}}&Input M & $183$& - &\\
&HLM1    & $700$& 128 800 & -\\
&HLM2   & $700$& 490 700 & -\\
&HLM3   & $350$& 245 350  &- \\
\hline
%&\multicolumn{4}{c}{Path A}\\
%\hline
\multirow{4}{*}{\rotatebox{90}{Path A}}&Input A  & $40{\times}9$ & - & $1{\times1}$\\
&ConvA1 & $25{\times}5{\times}5$ &650 & $1{\times}1$\\
&HLA1  & $700$& 3 150 000 &-\\
&HLA2  & $350$& 245 350 &-\\
\hline
%&\multicolumn{4}{c}{Shared layers}\\
%\hline
\multirow{3}{*}{\rotatebox{90}{Shared}}&HLS1 & $1600$& 3 681 600 &-\\
&HLS2 & $84$& 134 484 &-\\
&Output layer& $21$ & 1785 &-\\
% \hline
\end{tabular}\vspace*{-3pt}
\caption{\label{table:networkparameters} Hyper-parameters (for a single temporal scale)}\vspace*{-25pt}
\end{table}
\setlength{\tabcolsep}{1.4pt}

%\setlength{\tabcolsep}{4pt}
%\begin{table}[!t] \centering
%\caption{\label{table:networkparameters} Hyperparameters chosen for the deep learning models.}
%\vspace*{-2mm}
%\begin{tabular}{c|c|c|c}
%\hline\noalign{\smallskip}
%Layer & Parameters & Layer & Parameters \\
%\noalign{\smallskip}
%\hline
%\multicolumn{2}{c|}{Paths V1, V2} &  \multicolumn{2}{c}{Path M}\\
%%\aline
%Input D1,D2    & $36{\times}36{\times}5$             & Input M&   $183$\\
%Input C1,C2   & $36{\times}36{\times}5$      & HLM1                & $700$\\
%ConvD1 & $25{\times}5{\times}5{\times}3$ & HLM2               &$700$\\
%ConvD2 & $25{\times}5{\times}5$ & HLM3		    & $350$\\
%ConvC1 & $25{\times}5{\times}5{\times}3$ & \multicolumn{2}{c}{Path A}\\
%%\cline{3-4}
%ConvC2 & $25{\times}5{\times}5$& Input A& $40{\times}9$\\
%%\cline{3-4}
%Max pooling, steps & $2{\times}2{\times}3$& ConvA1 &$25{\times}5{\times}5$ \\
%HLV1  &  $900$ & HLA1 & $700$ \\
%HLV2  &  $450$ & HLA2 & $350$\\
%\noalign{\smallskip}
%\hline
%\noalign{\smallskip}
% \multicolumn{4}{c}{Shared layers}\\
% %\hline
% \noalign{\smallskip}
%\multicolumn{2}{c|}{Shared hidden layer HLS1} & \multicolumn{2}{c}{1600}\\
% \multicolumn{2}{c|}{Shared hidden layer HLS2} & \multicolumn{2}{c}{84}\\
% \multicolumn{2}{c|}{Number of outputs} & \multicolumn{2}{c}{21}\\
% \hline
%\end{tabular}
%\end{table}
%\setlength{\tabcolsep}{1.4pt}
%Neural, step 2 & $0.26$ \\
%Neural, step 3 & $1.02$\\
%Neural, step 4 & $2.20$\\
%Extremely randomized trees, step 4 & $1.00$\\
%Optimal weights for the model fusion are provided in Table \ref{table:weights}.
We followed the evaluation procedure proposed by the challenge organizers and adopted the Jaccard Index to quantify model performance:\vspace*{-5pt}
\begin{align}
  J_{s,n} = \dfrac{A_{s,n}\cap B_{s,n}}{A_{s,n} \cup B_{s,n}},
\end{align}\vspace*{-10pt}\\
where $A_{s,n}$ is the ground truth label of gesture $n$ in sequence $s$, and $B_{s,n}$ is the obtained prediction for the given gesture class in the same sequence. Here $A_{s,n}$ and $B_{s,n}$ are binary vectors where the frames in which the given gesture is being performed are marked with 1 and the rest with 0. Overall performance was calculated as the mean Jaccard index among all gesture categories and all sequences, with equal weight for all gesture classes. 

\subsection{Baseline models}
\label{baseline}
%\subsection{Depth and intensity video: extremely randomized trees}

%\begin{table}[t!] \centering
%\setlength{\tabcolsep}{4pt}
%\caption{\label{table:chalearn} \textit{ChaLearn 2014 ``Looking at people Challenge (track 3)''} results (top 10).}
%\vspace*{-2mm}
%\begin{tabular}{c|c|c||c|c|c}
%\hline\noalign{\smallskip}
%Rank & Team & Score & Rank & Team & Score\\
%\noalign{\smallskip}
%\hline
%\noalign{\smallskip}
% $\mathbf{1}$ & \textbf{Ours}             & $\mathbf{0.8500}$ & $6$  & stevenwudi \cite{chalearn6} & $0.7873$\\
% $2$ & CraSPN \cite{chalearn2}          & $0.8339$ & $7$  & ismar \cite{chalearn7}          & $0.7466$\\
% $3$ & JY \cite{chalearn3}                 & $0.8268$ & $8$  & Quads \cite{chalearn8}         & $0.7454$\\
% $4$ & CUHK-SWJTU \cite{chalearn4}& $0.7919$ & $9$  & Telepoints   & $0.6888$\\ 
% $5$ & lpigou \cite{chalearn5}           & $0.7888$ & $10$& TUM-fortiss \cite{chalearn10}& $0.6490$\\\hline
%\end{tabular}
%\setlength{\tabcolsep}{1.4pt}
%\end{table}

In addition to the main pipeline, we have implemented a baseline model based on an ensemble classifier trained in a similar iterative fashion but on purely handcrafted descriptors. 
The purpose of this comparison was to explore relative advantages and disadvantages of using learned representations as well as the nuances of fusion. 
We also 
% In addition, due to differences in feature formulation as well as in the nature of classifiers, we 
found it beneficial to combine the proposed deep network with the baseline method in a hybrid model 
%as separately two models make different errors 
(see Table \ref{table:results}).

%\subsubsection{Baseline visual models}
The baseline used for visual models is described in detail in~\cite{neverovaECCVW2014}.
We use depth and intensity hand images and extract three sets of features. HoG features describe the hand pose in the image plane. Histograms of depths describe pose along the third spatial dimension. The third set of features is comprised of derivatives of HOGs and depth histograms, which reflect temporal dynamics of hand shape.
% in each dynamic pose.

%\textbf{HoG features from intensity images.} First, we make use zero-mean and unit variance-normalized intensity images to extract HoG features $\mathbf{h}_{\textrm{int}}$~\cite{Dalal2005} at 9 orientations from a 2-level spatial pyramid~\cite{Lazebnik2006}, i.e.~from the whole image and a magnified version of it containing $3{\times}3$ cells.

%\textbf{Histograms of depths.}
%9-bin depth histograms %$\mathbf{h}_{\textrm{dep}}$ 
%are extracted on two scales from depth maps of both hands: from a whole map and from each quarter of an upsampled map.%version.% (by a factor of 2).

%\textbf{Derivatives of histograms.} First derivatives of HoGs and depth histograms are calculated: 
%$
%\delta\mathbf{h}(t){\approx}\mathbf{h}(t{+}1){-}\mathbf{h}(t{-}1),
%$
%where $\textbf{h}$ can stand for both $\textbf{h}_{\textrm{int}}$ and $\textbf{h}_{\textrm{dep}}$.
%Combined together, these three sets of features form a 270-dimensional descriptor %$[\mathbf{h}_{\textrm{int}}, \mathbf{h}_{\textrm{dep}}, \delta\mathbf{h}_{\textrm{int}}, \delta\mathbf{h}_{\textrm{dep}}]$ 
%
%for each frame and, consequently, a descriptor of dimension of 1350 for the dynamic pose of each hand.
%
%\textbf{Skeleton descriptor} remains the same as the one employed in the main model (see Section \ref{mocap}).

Extremely randomized trees (ERT)~\cite{geurts2006} are adopted for data fusion and gesture classification. %Ensemble methods of this sort have generally proven to be especially effective in conjunction with handcrafted features. 
%The ERT method differs from a classical random forest in the way how splits are computed. Instead of finding the exact optimal value of threshold for each subset of features, splitting rules are chosen from a set of randomly generated candidates. In our setting, the ERT was chosen due to $2\%$ higher performance in comparison with a random forest.
During training, we followed the same iterative strategy as in the case of the neural architecture (see \cite{neverovaECCVW2014} for more details).%. First, three ERT classifiers are trained independently on (i) skeleton descriptors (the same as described in Section \ref{sec:skeleton})), (ii) video features  for the right hand and (iii) video features for the left hand. Once training is completed, features from all modalities with importance above the mean value are selected and once again fused for training a new, general ERT classifier. Feature importance is calculated as mean decrease in impurity (i.e.~total decrease in node impurity weighted by proportion of samples reaching that node and averaged over all trees \cite{breiman1984}).

%At each step, ERT classifiers are trained with 300 estimators, an information gain criterion, no restrictions in depth and $\sqrt{N_f}$ features considered at each step (where $N_f$ is the total number of features). \vspace*{-5pt}
%For video stream: all features with importance higher than "mean" are used. Skeleton: 0.6*mean.

%\subsubsection{Baseline audio model}
A baseline has also been created for the audio channel, where we compare the proposed deep learning approach to a traditional phoneme recognition framework, as described in \cite{Neverova2013}, and implemented with the Julius engine \cite{bib:confinterspeechLeeKS01}. In this approach, each gesture is associated with a pre-defined vocabulary of possible ordered sequences of phonemes that can correspond to a single word or a phrase. After spotting and segmenting periods of voice activity, each utterance is assigned a n-best list of gesture classes with corresponding scores. Finally, frequencies of appearances of each gesture class in the list are treated as output class probabilities.

\subsection{Results on the ChaLearn 2014 LAP dataset}

\begin{table}[t!] \centering
\setlength{\tabcolsep}{4pt}
\begin{tabular}{c|c|c||c|c|c}
%\hline%\noalign{\smallskip}
\# & Team & Score & \# & Team & Score\\
%\noalign{\smallskip}
\hline
\noalign{\smallskip}
 $\mathbf{1}$ & \textbf{Ours \cite{neverovaECCVW2014}}           & $\mathbf{0.850}$ &$7$  & Camgoz et al. \cite{chalearn7}          & $0.747$ \\
 $2$ & \!Monnier\! et\! al.\! \cite{chalearn2}\!  & $0.834$ & $8$  & \!Evangelidis\! et\! al. \cite{chalearn8}\!         & $0.745$\\
 $3$ & Chang \cite{chalearn3}                 & $0.827$ &  $9$  & \!Undisclosed authors\! & $0.689$\\
 $4$ & Peng et al. \cite{chalearn4}& $0.792$ & $10$& Chen et al. \cite{chalearn10}& $0.649$\\ 
 $5$ & Pigou et al. \cite{chalearn5}           & $0.789$ & \multicolumn{3}{c}{$\ldots$}\\
 $6$  & Wu \cite{chalearn6} & $0.787$ & $17$  & Undisclosed authors & $0.271$\\
 \hline
 \noalign{\smallskip}
 \multicolumn{6}{c}{\textbf{Ours, improved results after the competition\quad $\mathbf{0.870}$}}\\
 \bottomrule
\end{tabular}
\setlength{\tabcolsep}{1.4pt}
\caption{\label{table:chalearn} \textit{Official ChaLearn 2014 LAP Challenge (track 3)} results, visual modalities only.}\vspace*{-15pt}
\end{table}
\begin{table}[!t] \centering
\setlength{\tabcolsep}{4pt}
\begin{tabular}{c|>{\centering\arraybackslash}m{1.cm}|>{\centering\arraybackslash}m{1.cm}|>{\centering\arraybackslash}m{1.6cm}|>{\centering\arraybackslash}m{1.cm}|>{\centering\arraybackslash}m{1.2cm}}
%\hline%\noalign{\smallskip}
Step & Pose& Video & Pose \& Video & Audio & All \\
%\noalign{\smallskip}
\hline
\noalign{\smallskip}
 $2$ & $0.823$ & $0.818$ & $0.856$ & $0.709$ & $0.870$\\%$0.853$\\
 $3$ & $0.824$ & $0.817$ & $0.859$ & $0.731$ & $0.873$\\%$0.864$\\
 $4$ & $0.827$ & $0.825$ & $0.859$ & $0.714$ & $0.880$\\\hline%$0.866$\\\hline
 \noalign{\smallskip}
all    & ${0.831}$ & ${0.836}$ & $\textbf{0.868}$ & {0.734} & $\textbf{0.881}$%\\\bottomrule
\end{tabular}
\setlength{\tabcolsep}{1.4pt}
\caption{\label{table:neuralresults} Post-competition performance at different temporal scales with gesture localization (Jaccard index).}\vspace*{-25pt}
\end{table}

%\begin{table}[!t] \centering
%\setlength{\tabcolsep}{4pt}
%\caption{\label{table:neuralresults}
%\nn{New version}Post-competition performance at different temporal scales (deep learning + binary motion detector). All numbers reported in the table are the Jaccard Index.}
%\vspace*{-2mm}
%\begin{tabular}{c|>{\centering\arraybackslash}m{1.cm}|>{\centering\arraybackslash}m{1.cm}|>{\centering\arraybackslash}m{1.6cm}|>{\centering\arraybackslash}m{1.cm}|>{\centering\arraybackslash}m{1.2cm}}
%\hline\noalign{\smallskip}
%Step & Pose& Video & Pose \& Video & Audio & All \nn{no}\\
%\noalign{\smallskip}
%\hline
%\noalign{\smallskip}
% $2$ & $0.823$ & $0.818$ & $0.000$ & $0.709$ & $0.000$\\
% $3$ & $0.824$ & $0.817$ & $0.000$ & $0.731$ & $0.000$\\
% $4$ & $0.827$ & $0.825$ & $0.000$ & $0.714$ & $0.876$\\\hline
% \noalign{\smallskip}
%all    & ${0.831}$ & ${0.836}$ & $\textbf{0.000}$ & {0.734} & $\textbf{0.000}$\\\bottomrule
%\end{tabular}
%\setlength{\tabcolsep}{1.4pt}
%\end{table}

The top 10 scores of the ChaLearn 2014 LAP Challenge (track 3) are reported in Table~\ref{table:chalearn}. Our winning entry \cite{neverovaECCVW2014} corresponding to a hybrid model (i.e.~a combination of the proposed deep neural architecture and the ERT baseline model) surpasses the second best score by a margin of 1.61 percentage points. We also note that the multi-scale neural architecture still achieves the best performance, as well as the top one-scale neural model alone (see Tables \ref{table:neuralresults} and \ref{table:results}).
In post-challenge work we were able to further improve the score by 2.0 percentage points to 0.870 by introducing additional capacity into the model, optimizing the architectures of the video and skeleton paths and employing a more advanced training and fusion procedure (ModDrop) which was not used for the challenge submission.

Detailed information on the performance of the neural architectures for each modality and at each scale is provided in  Table~\ref{table:neuralresults}, including both the multi-modal setting and per-modality tests.
%While in our previous work \cite{neverovaECCVW2014} we observed significant degradation in performance of the mocap model in the case of dense temporal sampling (${s=}2$), 
Our experiments have proven that useful information can be extracted at any scale given sufficient model capacity (which is typically higher for small temporal steps). Trained independently, articulated pose models corresponding to different temporal scales demonstrate similar performance if predictions are refined by the gesture localization module. 
Video streams, containing information about hand shape and articulation, are also insensitive to the sampling step and demonstrate good performance even for short spatio-temporal blocks. 
%This signifies that in the context of this dataset, a body pose is interesting exclusively in terms of its dynamics, while hand postures are fairly discriminative alone, even in nearly static mode. 
The overall highest score is nevertheless obtained in the case of a dynamic pose with duration roughly corresponding to the length of an average gesture ($s{=}4$, i.e. covers the time period of 17 frames).

Table \ref{table:mocapresults} illustrates performance of the proposed modality-specific architectures compared to results reported by other participants of the challenge. For both visual channels: articulated pose and video, our method significantly outperforms the proposed alternatives.

%%%%
\setlength{\tabcolsep}{4pt}
\begin{table}[!t] \centering
\begin{minipage}[c]{\linewidth}\centering
\begin{tabular}{l|c|c}
%\hline%\noalign{\smallskip}
\hspace*{\fill} Model \hspace*{\fill} & Pose (mocap) & Video\\% & Subset\\
%\noalign{\smallskip}
\hline
\noalign{\smallskip}
Evangelidis\! et\! al. \cite{chalearn8}, submitted entry  & $0.745$ & --\\
Camgoz et al. \cite{chalearn7}      & $0.747$ & --\\%  & test\\
Evangelidis\! et\! al. \cite{chalearn8}, after competition  &$0.768$ & --\\%  & test\\
Wu and Shao \cite{chalearn6}        & $0.787$ & $0.637$\\%  & test\\
Monnier et al. \cite{chalearn2} (validation set)       & $0.791$& --\\%  & validation\\
Chang \cite{chalearn3}   & $0.795$ & --\\%  & test\\
Pigou et al. \cite{chalearn5}  & -- &$0.789$\\%  & test\\
Peng et al. \cite{chalearn4}  & -- & $0.792$\\%  & test\\
Ours, submitted entry \cite{neverovaECCVW2014}  & $\mathbf{0.808}$ & $\mathbf{0.810}$\\% & test\\
Ours, after competition  & $\mathbf{0.831}$ & $\mathbf{0.836}$\\%  & test\\
%\bottomrule
\end{tabular}%\vspace*{-10pt}
%\footnotetext[1]{Reported on the validation set}
\end{minipage}
\caption{\label{table:mocapresults}Official ChaLearn 2014 LAP Challenge
results on mocap and video data (Jaccard index).}\vspace*{-12pt}
\end{table}
\setlength{\tabcolsep}{1.4pt}

\setlength{\tabcolsep}{4pt}
\begin{table}[!t] \centering
\vspace*{-2mm}
\begin{tabular}{l|>{\centering\arraybackslash}m{1.5cm}|>{\centering\arraybackslash}m{1.5cm}|>{\centering\arraybackslash}m{1cm}}
%\hline%\noalign{\smallskip}
\hspace*{\fill} Model \hspace*{\fill} & Without localization & With localization & (Virtual) rank\\
%\noalign{\smallskip}
\hline
%\noalign{\smallskip}
ERT (baseline)              & \centering$0.729$ & \centering$0.781$ & (6)\\
Ours \cite{neverovaECCVW2014} & \centering$0.812$ & \centering${0.849}$ & (1)\\
Ours \cite{neverovaECCVW2014}  + ERT & \centering$0.814$ & \centering${0.850}$ & 1\\
Ours (improved) & \centering$0.821$ & \centering$\textbf{0.868}$ & (1)\\
Ours (improved) + ERT  & \centering$0.829$ & \centering$\textbf{0.870}$ & (1)\\
%\bottomrule
\end{tabular}
\caption{\label{table:results}Performance on visual modalities (Jaccard Index).}\vspace*{-25pt}
\end{table}
\setlength{\tabcolsep}{1.4pt}

The comparative performance of the baseline and hybrid models for visual modalities are reported in Table~\ref{table:results}.
In spite of the low scores of the isolated ERT baseline model, fusing its outputs with those provided by the neural architecture is still slightly beneficial, mostly due to differences in feature formulation in the video channel (adding ERT to mocap alone did not result in a significant gain).

For each combination, we provide results obtained with a classification module alone (without additional gesture localization) and coupled with the binary motion detector.
 The experiments demonstrate that the localization module contributes significantly to overall performance.

\begin{table}[t]
\begin{center}
\begin{tabular}{c|>{\centering\arraybackslash}m{30pt}|>{\centering\arraybackslash}m{35pt}|>{\centering\arraybackslash}m{40pt}|>{\centering\arraybackslash}m{30pt}}
%\arrayrulecolor{cwblue1}
 %\toprule
\hspace{\fill} Method \hspace{\fill}  & Recall, \% & Precision, \% & F-measure, \% &Jaccard index\\
\midrule
Phoneme recognition \cite{Neverova2013}& $64.70$ 	&$50.11$ 	&$56.50$ & $0.256$\\
Learned representation & $87.42$ 	&$73.34$ 	&$79.71$ & $0.545$\\
%\bottomrule
\end{tabular}
\end{center}\vspace*{-5pt}
\caption{\label{table:audioresults} Comparison of proposed and baseline approaches to gesture recognition from audio.\vspace*{-25pt}}%: a method based on phoneme recognition (proposed in our earlier work \cite{Neverova2013}) and the representation learning pipeline presented in Sec.~\ref{sec:audio}. } %Curly brackets signify iterative fusion of modalities one by one, such that the notation \{Depth, Pose, Audio\}, for example, means that two depth channels (for the right and for the left hands) have been fused first with following integrating of the articulated pose channel and finally of the audio path. }
\end{table}

\subsection{Results on the ChaLearn 2014 LAP dataset augmented with audio}%Results on the extended dataset including audio}

To demonstrate how the proposed model can be further extended with arbitrary data modalities, we introduce speech to the existing setup. In this setting, each gesture in the dataset is accompanied by a word or a short phrase expressing the same meaning and pronounced by each actor while performing the gesture. 
As expected, introducing a new data channel resulted in significant gain in classification performance (1.3 points on the Jaccard index, see Table  \ref{table:neuralresults}).

As with the other modalities, an audio-specific neural network was first pretrained discriminatively on the audio data alone. Next, the same fusion procedure was employed without any change.
In this case, the quality of predictions produced by the audio path depends on the temporal sampling frequency: the best performance was achieved for dynamic poses of duration ${\sim}0.5\,\textrm{s}$ (see Table  \ref{table:neuralresults}). %The same table also illustrates the contribution of the audio channel in the overall classification performance. 

Although the overall score after adding the speech channel is improved significantly, the audio modality alone does not perform so well.
This can be partly explained by natural gesture-speech desynchronisation resulting in poor audio-based gesture localization. In this dataset, gestures are annotated based on video recordings, while pronounced words and phrases are typically shorter in time than movements. Moreover, depending on the style of each actor, vocalisation can be either slightly delayed to coincide with gesture culmination, or can be slightly ahead of time announcing the gesture. Therefore, the audio signal alone does not allow the model to robustly predict the start- and end-points of a gesture, which results in poor Jaccard scores.

%\subsection{Impact of the audio channel}

Table \ref{table:audioresults} compares the performance of the proposed solution based on learning representations from mel-frequency spectrograms with the baseline model involving traditional phoneme recognition \cite{Neverova2013}. Here, we report the values of Jaccard indices for the reference, but, as it was mentioned above, accurate gesture localization based exclusively on the audio channel is not possible for reasons outside of the model's control. To make a more meaningful comparison of the classification performance, we report recall, precision and F-measure for each model. In this case we assume that the gesture was correctly detected and recognised if temporal overlap between predicted and ground truth gestures is at least 20\%.

Our results show that, in the given context, employing the deep learning approach drastically improves performance in comparison with the traditional framework based on phoneme recognition.

\subsection{Impact of the different fusion strategies}

We explore the relative advantages of different training strategies, starting 
%, including such aspects as pretraining, initialization of shared layers and general and modality-specific regularization (dropout, ModDrop).
with preliminary experiments on the MNIST dataset \cite{mnist} and then a more extensive analysis on the ChaLearn 2014 dataset augmented with audio.

\subsubsection{Preliminary experiments on MNIST dataset}

As a sanity check of ModDrop fusion, we transform the MNIST dataset \cite{mnist} to imitate multi-modal data. 
A classic deep learning benchmark, MNIST consists of $28{\times}28$ grayscale images of handwritten digits, where 60k examples are used for training and 10k images are used for testing. We use the original version with no data augmentation. We also avoid any data preprocessing and apply a simple architecture: a multi-layer perceptron with two hidden layers (i.e.~no convolutional layers).

\begin{figure}[!t]
\centering
\includegraphics[height=45mm]{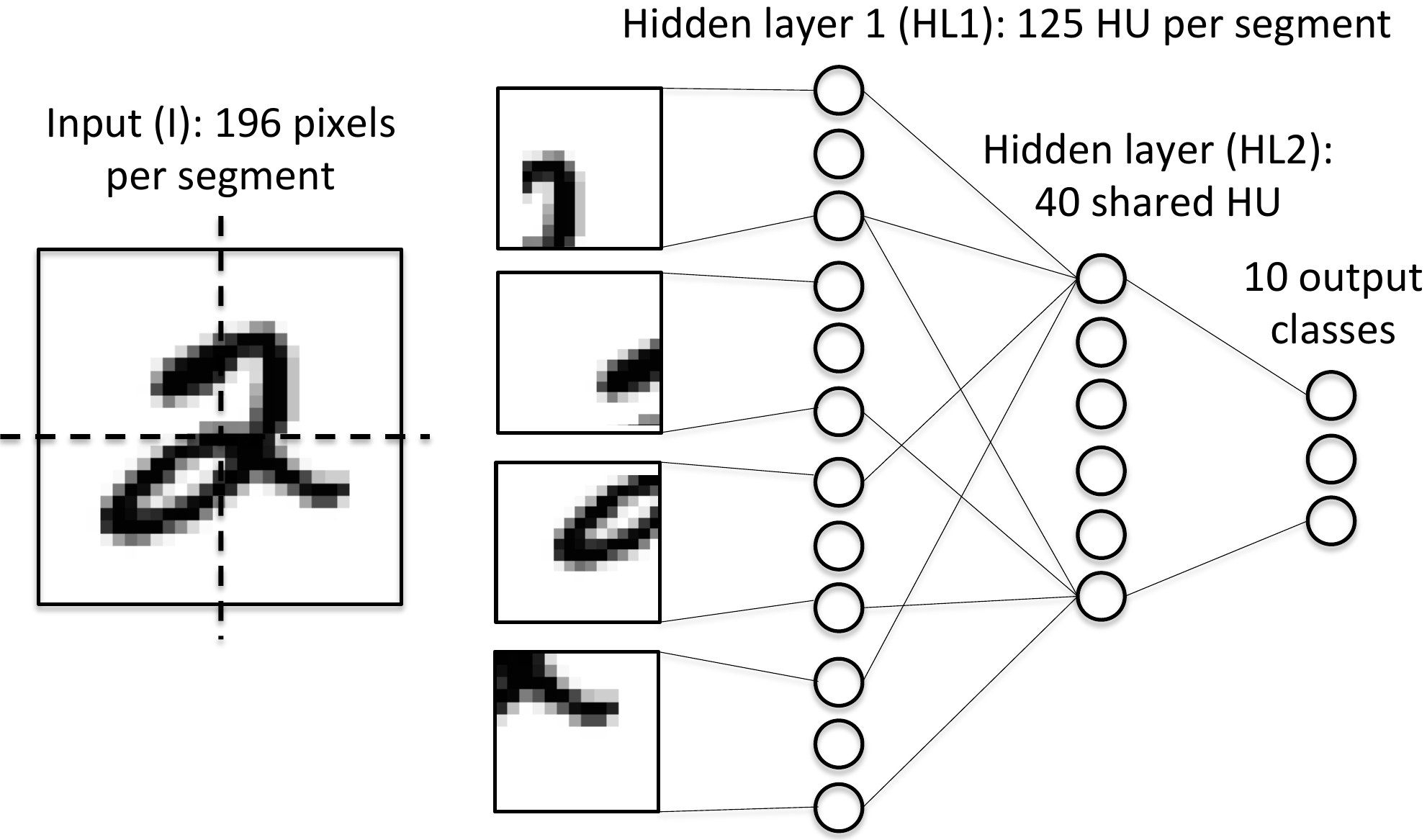}\vspace*{-5pt}
\caption{``Multi-modal" setting for the MNIST dataset.}% Each digit is divided into 4 non-overlapping segments and corresponding subsets of modality-specific units HL1 are first discriminatively pretrained independently.}%on a corresponding  image segment independently. The fusion is performed with a shared layer HL2 and the output layer.}
\label{fig:mnistfigure}\vspace*{-10pt}
\end{figure}

%To work out the idea of multi-modal fusion, 
We cut each digit image into 4 quarters and assume that each quarter corresponds to one modality (see Fig.~\ref{fig:mnistfigure}).
In spite of the apparent simplicity of this formulation, we show that the obtained results accurately reflect the dynamics of a real multi-modal setup.

The multi-signal training objective is two-fold: first, we optimize the architecture and the training procedure to obtain the best overall performance on the full set of modalities. The second goal is to make the model robust to missing signals or a high level of noise in the separate channels. To explore the latter aspect, during test time we occlude one or more image quarters or add pepper noise to one or more image parts.

\begin{table}[t]
\begin{center}
\begin{tabular}{c|c|c|c|c|r}
%\toprule
%\multicolumn{6}{c}{Fully connected setting}\\[3pt]\hline
\multicolumn{3}{c|}{Training mode} & \hspace*{0pt} Errors & \multicolumn{2}{c}{\# of parameters}\\
\midrule
\multicolumn{3}{c|}{Dropout, 784-1200-1200-10 \cite{dropout}} & 107 & 2395210 & N\\
\multicolumn{3}{c|}{Dropout, 784-500-40-10 (ours)} & 119 & 412950 &0.17N \\[1pt]
\bottomrule
\multicolumn{6}{c}{\vphantom{${a^2}^2$}(a) Fully connected setting}\\[6pt]
%\toprule
%\multicolumn{6}{c}{"Multi-modal" setting, $196{\times}4$-$125{\times}4$-40-10}\\
%\midrule
Pretraining (HL1) & Dropout (I) & ModDrop (I) & Errors & \multicolumn{2}{c}{\# of parameters} \\
\hline
no & no & no & 142 & \multirow{5}{*}{118950} & \multirow{5}{*}{$0.05N$}\\
no & \textbf{yes} & no & 123 & &\\
\textbf{yes} & no & no & 118 & &\\
\textbf{yes} & \textbf{yes} & no & \textbf{102} & &\\
\textbf{yes} & \textbf{yes} & \textbf{yes} & \textbf{102} & &\\[1pt]
\bottomrule 
\multicolumn{6}{c}{\vphantom{${a^2}^2$}(b) ``Multi-modal" setting, $196{\times}4$-$125{\times}4$-40-10}
\end{tabular}
\end{center}\vspace*{-8pt}
\caption{Experiments on the MNIST dataset.}%: (a) scores of fully connected single-modal architecture and a more compact network used as a starting point for our experiments; (b) In the multi-modal setting, the first hidden layer is split between 4 channels, which results in a tree-structured architecture.}%$196{\times}4$(I)-$125{\times}4$(HL1)-40(HL2)-10 
\label{table:mnisttraining}\vspace*{-25pt}
\end{table}

\begin{table}[t]
\begin{center}
\begin{tabular}{l|c|c}
%\arrayrulecolor{cwblue1} 
%\toprule
\hspace{\fill} Training mode \hspace{\fill}  &\hspace{3pt}Dropout\hspace{3pt}  & \hspace{3pt}Dropout + ModDrop\\
\noalign{\smallskip}
\hline
%\noalign{\smallskip}
\multicolumn{3}{c}{Missing segments, test error, \%}\\\hline
All segments visible & 1.02 & 1.02 \\
1 segment covered & 10.74 & 2.30 \\
2 segments covered & 35.91 & 7.19\\
3 segments covered & 68.03 & 24.88\\\hline
\multicolumn{3}{c}{Pepper noise 50\% on segments, test error, \%}\\\hline
All clean& 1.02 & 1.02 \\
1 corrupted segment & 1.74 & 1.56 \\
2 corrupted segments & 2.93 & 2.43\\
3 corrupted segments & 4.37 & 3.56\\
All segments corrupted& 7.27 & 6.42\\
%\bottomrule
\end{tabular}
\caption{Effect of ModDrop training under occlusion and noise.}%: the network is more robust to missing signals and noise in separate channels.}
\label{table:mnistdrop}
\end{center}\vspace*{-25pt}
\end{table}
%%%%%%

Currently, the state-of-the-art for a fully-connected 782-1200-1200-10 network with dropout regularization (50\% for hidden units and 20\% for the input) and tanh activations \cite{dropout} is 107 errors on the MNIST test set 
(see Table \ref{table:mnisttraining}). 
In this case, the number of units in the hidden layer is unnecessarily large, which is exploited by dropout-like strategies. When real-time performance is a constraint, this redundancy in the number of operations becomes a serious limitation. Instead, switching to our tree-structured network (i.e.~a network with separated modality-specific input layers connected to a set of shared layers) is helpful for independent modality-wise tuning of model capacity, which in this case does not have to be uniformly distributed over the input units.
% and minimizing the redundancy
For this multi-modal setting we optimized the number of units (125) for each channel and do not apply dropout to the hidden units (which in this case turns out to be harmful due to the compactness of the model), limiting ourselves to dropping out the inputs at a rate of 20\%. In addition, we apply ModDrop on the input, where the probability of each segment to be dropped is 10\%.

The results in Table~\ref{table:mnisttraining} show that 
%\begin{itemize}
%\item 
separate pretraining of modality-specific paths generally yields better performance and leads to a significant decrease in the number of parameters due to the capacity restriction placed on each channel. This is apparent in the 4\textsuperscript{th} row of Table~\ref{table:mnisttraining}b: with pretraining, better performance (102 errors) is obtained with 20 times less parameters.

MNIST results under occlusion and noise are presented in Table~\ref{table:mnistdrop}. We see that ModDrop, while not affecting the overall performance on MNIST, makes the model significantly less sensitive to occlusion and noise. 
\subsubsection{Experiments on ChaLearn 2014 LAP with audio}
%\nn{Rewrite completely.}

\begin{table}[t]
\begin{center}
\begin{tabular}{c|c|c|c|c}
%\arrayrulecolor{cwblue1} 
%\toprule
\phantom{m}Pretraining\phantom{m} & \phantom{m}Dropout\phantom{m} & \phantom{m}Initial.\phantom{m} & \phantom{m}ModDrop\phantom{m} & \phantom{m}Accuracy, \%\phantom{m}\\[1pt]
\hline
no & no & no & no & $91.94$\\
no & \textbf{yes} & no & no & $93.33$\\
\textbf{yes} & no & no & no & $94.96$\\
\textbf{yes} & \textbf{yes} & no & no & $96.31$\\
\textbf{yes} & \textbf{yes} & \textbf{yes} & no & $96.77$\\
\textbf{yes} & \textbf{yes} & \textbf{yes} & \textbf{yes} & $\mathbf{96.81}$\\ 
%\bottomrule
\end{tabular}
\end{center}\vspace*{-5pt}
\caption{ Comparison of different training strategies on the ChaLearn 2014  LAP dataset augmented with audio.}% "Pretraining" column corresponds to modality-specific paths while "Initial." indicates whether or not the shared layers have also been pre-initialized with pre-trained diagonal blocks. In all cases, dropout (20\%) and ModDrop (probability 0.1) are applied to the input signal.  The accuracy corresponds to per-block classification on the validation set.}
\label{table:strategy}\vspace*{-20pt}
\end{table}

In a real multi-modal setting, optimizing and balancing a tree-structured architecture is an extremely difficult task as its separated parallel paths vary in complexity and operate on different feature spaces. The problem becomes even harder under the constraint of real-time performance and, consequently, the limited capacity of the network.

Our experiments have shown that insufficient modelling capacity of one of the modality-specific subnetworks leads to a drastic degradation in performance of the whole system due to the multiplicative nature of the fusion process. Those bottlenecks are typically difficult to find without thorough per-channel testing.

%For the proposed solution, contribution of each modality into the final predictions, their groupings and ordering of fusion steps are graphically shown in Fig. \ref{fig:error}. Values along the abscissa correspond to a number of a training stage: thus, fusion of depth and intensity channels (point (3)) is followed by combining of streams for two hands (point (4)) and iterative integration of mocap (point (6)) and audio (point (8)) signals. The final step consisting in weighted averaging of predictions across scales correspond to point (9). Points (1), (2), (5) and (7) illustrate validation classification errors for isolaited data channels.

%\begin{figure}[!t]
%\centering
%\includegraphics[width=\linewidth]{error.pdf}
%\caption{\nn{Update values. Flip?} Evolution of classification error on each step of the iterative fusion process in the tree-structured deep neural architecture}
%\label{fig:error}
%\end{figure}

We propose to start by optimizing the architecture and hyper-parameters for each modality separately through discriminative pretraining. During fusion, input paths are initialized with pretrained values and fine-tuned while training the output shared layers.

Furthermore, the shared layers can also be initialized with pretrained diagonal blocks as described in Section~\ref{sec:traininginitialization}, which results in a
 significant speed up in the training process. We have observed that in this case, setting the biases of the shared hidden layer is critical in converging to a better solution. 

As in the case of the MNIST experiments, we apply 20\% dropout on the input signal and ModDrop with probability of 10\% (optimized on the validation set). As before, dropping hidden units during training led to degradation in performance of our architecture due to its compactness.

A comparative analysis of the efficiency of various training strategies is reported in Table \ref{table:strategy}. Here, we provide validation error of per dynamic pose classification as a direct indicator of convergence of training. The ``Pretraining'' column corresponds to modality-specific paths while ``Initial.'' indicates whether or not the shared layers have also been pre-initialized with pretrained diagonal blocks. In all cases, dropout (20\%) and ModDrop (10\%) are applied to the input signal.  Accuracy corresponds to per-block classification on the validation set.

Differences in effectiveness of different strategies agree well with what we have observed previously on MNIST. Modality-wise pretraining and regularization of the input have a strong positive effect on performance. Interestingly, in this case ModDrop resulted in further improvement in scores even for the complete set of modalities (while increasing the dropout rate did not have the same effect).

Analysis of the network behaviour in conditions of noisy or missing signals in one or several channels is provided in Table \ref{table:strategymoddrop}. Once again, ModDrop regularization resulted in much better network stability with respect to signal corruption and loss.

\begin{table}[t]
\begin{center}
\begin{tabular}{l|c|c|c|c}
%\arrayrulecolor{cwblue1} 
%\toprule
\hspace{\fill} Modality \hspace{\fill}  & \multicolumn{2}{c|}{Dropout} & \multicolumn{2}{c}{Dropout + ModDrop}\\
%\noalign{\smallskip}
\hline
& Accuracy, \% & Jaccard index & Accuracy, \% & Jaccard index\\
\hline
All present & 96.77 & 0.876 & 96.81 & 0.880\\
\hline
\multicolumn{5}{c}{Missing signals in separate channels}\\
\hline
%Left hand, color & 6.73 & XXX & 4.77 & XXX\\
%Left hand, depth & 7.33 & XXX & 5.79 & XXX \\
%Right hand, color & 11.63 & XXX & 5.94 & XXX\\
%Right hand, depth & 11.29 & XXX & 6.90 & XXX\\ 
Left hand & 89.09 & 0.826 & 91.87 & 0.832 \\
Right hand & 81.25 & 0.740 & 85.36 & 0.796 \\
Both hands & 53.13 & 0.466 & 73.28 & 0.680\\
Mocap & 38.41 & 0.306 & 92.82 & 0.859 \\
Audio & 84.10 & 0.789 & 92.59 & 0.854\\
\hline
\multicolumn{5}{c}{Pepper noise 50\% in channels}\\
\hline
%Left hand, color & 3.84 & XXX & XXX & XXX \\
%Left hand, depth & 3.84 & XXX & XXX & XXX \\
%Right hand, color & 3.76 & XXX & XXX & XXX\\
%Right hand, depth & 3.73 & XXX & XXX & XXX\\ 
Left hand & 95.36 & 0.874 & 95.75 & 0.874\\ % nofixedbias, other 4.64
Right hand & 95.48 & 0.873 & 95.92 & 0.874\\ % nofixedbias, other 4.52
Both hands & 94.55 & 0.872 & 95.06 & 0.875\\ % nofixedbias, other 5.45
Mocap & 93.31 & 0.867 & 94.28 & 0.878\\ %resumed_nofixedbias, other 6.01
Audio & 94.76 & 0.867 & 94.96 & 0.872\\ %resumed_nofixedbias, other 5.18
%\bottomrule
\end{tabular}
\end{center}\vspace*{-5pt}
\caption{Effect of ModDrop on ChaLearn 2014+audio.}% As before, ModDrop training results in more stable output predictions even in conditions of missing or corrupted signals. Furthermore, In this case of real multi-modal setting it also improves model performance on a complete set of modalities.}
\label{table:strategymoddrop}\vspace*{-25pt}
\end{table}

\section{Conclusion}

We have described a generalized method for gesture and near-range
action recognition from a combination of range video data and
articulated pose. Each of the visual modalities captures spatial
information at a particular spatial scale (such as motion of the upper body or a hand), and the whole system operates at two temporal scales. 

The model can be further extended and augmented with arbitrary
channels (depending on available sensors) by introducing additional
parallel pathways without significant changes in the general
structure. We illustrate this concept by augmenting video with speech. Multiple spatial and temporal scales per channel can easily
be integrated. 

Finally, we have explored various aspects of multi-modal fusion in terms of joint performance on a complete set of modalities as well as robustness of the classifier with respect to noise and dropping of one or several data channels. As a result, we have proposed a modality-wise regularisation strategy (ModDrop) allowing our model to obtain stable predictions even when inputs are corrupted.

%As future work, we aim to reformulate and generalize the problem from gesture detection and recognition to sensing gesture parameters (e.g.~motion amplitude in scrolling-like gestures). In this case, the video stream providing information about subtle hand movements will play a primary role.
%Another interesting direction would be a deeper exploration into the
%dynamics of cross-modality dependencies and considering full signal reconstruction (similar to \cite{Ngiam2011}) in the case when one or more modality data inputs are missing.
%Recent advances in multimodal deep learning \citep{Ngiam2011}, multimodal DBMs \citep{Srivastava2013}.
%Learning to reconstruct one modality from another.

\subsubsection*{Acknowledgement}
This work has been partly financed through the French grant Interabot,
a project of type ``Investissement's d'Avenir / Briques G\'en\'eriques
du Logiciel Embarqu\'e''.

\vskip 0.2in
\bibliographystyle{IEEEtran}
%\bibliography{iccv2013ws}

\bibliography{egbib}

\begin{thebibliography}{10}
\providecommand{\url}[1]{#1}
\csname url@samestyle\endcsname
\providecommand{\newblock}{\relax}
\providecommand{\bibinfo}[2]{#2}
\providecommand{\BIBentrySTDinterwordspacing}{\spaceskip=0pt\relax}
\providecommand{\BIBentryALTinterwordstretchfactor}{4}
\providecommand{\BIBentryALTinterwordspacing}{\spaceskip=\fontdimen2\font plus
\BIBentryALTinterwordstretchfactor\fontdimen3\font minus
  \fontdimen4\font\relax}
\providecommand{\BIBforeignlanguage}[2]{{%
\expandafter\ifx\csname l@#1\endcsname\relax
\typeout{** WARNING: IEEEtran.bst: No hyphenation pattern has been}%
\typeout{** loaded for the language `#1'. Using the pattern for}%
\typeout{** the default language instead.}%
\else
\language=\csname l@#1\endcsname
\fi
#2}}
\providecommand{\BIBdecl}{\relax}
\BIBdecl

\bibitem{Girshick2014}
R.~Girshick, J.~Donahue, T.~Darrell, and J.~Malik, ``{Rich feature hierarchies
  for accurate object detection and semantic segmentation},'' in \emph{CVPR},
  2014.

\bibitem{Sermanet2014}
P.~Sermanet, D.~Eigen, X.~Zhang, M.~Mathieu, R.~Fergus, and Y.~LeCun,
  ``{OverFeat: Integrated Recognition, Localization and Detection using
  Convolutional Networks},'' in \emph{ICLR}, 2014.

\bibitem{Krizhevsky2012}
A.~Krizhevsky, I.~Sutskever, and G.~Hinton, ``{ImageNet Classification with
  Deep Convolutional Neural Networks},'' in \emph{NIPS}, 2012.

\bibitem{Farabet2013}
C.~Farabet, C.~Couprie, L.~Najman, and Y.~LeCun, ``{Learning Hierarchical
  Features for Scene Labeling},'' in \emph{PAMI}, 2013.

\bibitem{Couprie2014}
C.~Couprie, F.~Clément, L.~Najman, and Y.~LeCun, ``{Indoor Semantic
  Segmentation using depth information},'' in \emph{ICLR}, 2014.

\bibitem{LeCun1998}
Y.~LeCun, L.~Bottou, Y.~Bengio, and P.~Haffner, ``{Gradient-based learning
  applied to document recognition},'' \emph{Proceedings of the IEEE}, vol.~86,
  no.~11, pp. 2278--2324, 1998.

\bibitem{Kahou2013}
S.~E. Kahou \emph{et~al.}, ``{Combining modality specific deep neural networks
  for emotion recognition in video},'' in \emph{ICMI}, 2013.

\bibitem{Taigman2014}
Y.~Taigman, M.~Yang, M.~A. Ranzato, and L.~Wolf, ``{DeepFace: Closing the Gap
  to Human-Level Performance in Face Verification},'' in \emph{CVPR}, 2014.

\bibitem{Baccouche2012}
M.~Baccouche, F.~Mamalet, C.~Wolf, C.~Garcia, and A.~Baskurt,
  ``{Spatio-Temporal Convolutional Sparse Auto-Encoder for Sequence
  Classification},'' in \emph{BMVC}, 2012.

\bibitem{Karpathy2014}
A.~Karpathy, G.~Toderici, S.~Shetty, T.~Leung, R.~Sukthankar, and F.-F. Li,
  ``{Large-scale Video Classification with Convolutional Neural Networks},'' in
  \emph{CVPR}, 2014.

\bibitem{Simonyan2014}
K.\hspace{2pt}Simonyan and \!A.\hspace{2pt}Zisserman,
  ``{Two-Stream\hspace{-1pt} Convolutional\hspace{-1pt} Networks\hspace{-1pt}
  for\hspace*{-1pt} Action\hspace{-1pt} Recognition\hspace{-1pt}
  in\hspace{-1pt} Videos},'' in \emph{\!arXiv:1406.2199v1}, 2014.

\bibitem{modeep}
A.~Jain, J.~Tompson, Y.~LeCun, and C.~Bregler, ``{MoDeep: A Deep Learning
  Framework Using Motion Features for Human Pose Estimation},'' in \emph{ACCV},
  2014.

\bibitem{zhou2013}
F.~Zhou, F.~De~la Torre, and J.-K. Hodgins, ``{Hierarchical Aligned Cluster
  Analysis for Temporal Clustering of Human Motion},'' in \emph{PAMI}, 2013.

\bibitem{escalera2014chalearn}
S.\hspace{1pt}Escalera, X.\hspace{1pt}Bar\'{o}, J.\hspace{1pt}Gonz\`{a}lez,
  M.\hspace{1pt}Bautista, M.\hspace{1pt}Madadi, M.\hspace{1pt}Reyes,
  V.\hspace{1pt}Ponce, H.\hspace{1pt}Escalante, J.\hspace{1pt}Shotton, and
  I.\hspace{1pt}Guyon, ``{ChaLearn Looking at People Challenge 2014: Dataset
  and Results},'' in \emph{ECCVW}, 2014.

\bibitem{Wang2013}
H.~Wang, A.~Kl\"{a}ser, C.~Schmid, and C.-L. Liu, ``{Dense trajectories and
  motion boundary descriptors for action recognition},'' \emph{IJCV}, 2013.

\bibitem{Shotton2011}
J.~Shotton, A.~Fitzgibbon, M.~Cook, T.~Sharp, M.~Finocchio, R.~Moore,
  A.~Kipman, and A.~Blake, ``{Real-time human pose recognition in parts from
  single depth images},'' in \emph{CVPR}, 2011.

\bibitem{Keskin2011}
C.~Keskin, F.~Kira\c{c}, Y.~Kara, and L.~Akarun, ``{Real time hand pose
  estimation using depth sensors},'' in \emph{ICCV Workshop}, 2011.

\bibitem{Tang2013}
D.~Tang, T.-H. Yu, and T.-K. Kim, ``{Real-time Articulated Hand Pose Estimation
  using Semi-supervised Transductive Regression Forests},'' in \emph{ICCV},
  2013.

\bibitem{Tang2014}
D.~Tang, H.~J. Chang, A.~Tejani, and T.-K. Kim, ``{Latent Regression Forest:
  Structured Estimation of 3D Articulated Hand Posture},'' in \emph{CVPR},
  2014.

\bibitem{Tompson2014}
J.~Tompson, M.~Stein, Y.~LeCun, and K.~Perlin, ``{Real-Time Continuous Pose
  Recovery of Human Hands Using Convolutional Networks},'' in \emph{ACM
  Transaction on Graphics}, 2014.

\bibitem{neverovaACCV2014}
N.~Neverova, C.~Wolf, G.~Taylor, and F.~Nebout, ``{Hand segmentation with
  structured convolutional learning},'' in \emph{ACCV}, 2014.

\bibitem{Oikonomidis2011}
I.~Oikonomidis, N.~Kyriazis, and A.~Argyros, ``{Efficient model-based 3D
  tracking of hand articulations using Kinect},'' in \emph{BMVC}, 2011.

\bibitem{Qian2014}
C.~Qian, X.~Sun, Y.~Wei, X.~Tang, and J.~Sun, ``{Realtime and Robust Hand
  Tracking from Depth},'' in \emph{CVPR}, 2014.

\bibitem{Wang2013a}
F.~Wang and Y.~Li, ``{Beyond Physical Connections: Tree Models in Human Pose
  Estimation},'' in \emph{CVPR}, 2013.

\bibitem{chen2014}
X.~Chen, R.~Mottaghi, X.~Liu, S.~Fidler, R.~Urtasun, and A.~Yuille, ``{Detect
  What You Can: Detecting and Representing Objects using Holistic Models and
  Body Parts},'' in \emph{CVPR}, 2014.

\bibitem{Wang2012}
J.~Wang, Z.~Liu, Y.~Wu, and J.~Yuan, ``{Mining actionlet ensemble for action
  recognition with depth cameras},'' in \emph{CVPR}, 2012.

\bibitem{Sung2012}
J.~Sung, C.~Ponce, B.~Selman, and A.~Saxena, ``{Unstructured Human Activity
  Detection from RGBD Images},'' in \emph{ICRA}, 2012.

\bibitem{Chen2013}
X.~Chen and M.~Koskela, ``{Online RGB-D gesture recognition with extreme
  learning machines},'' in \emph{ICMI}, 2013.

\bibitem{chalearn2}
C.~Monnier, S.~German, and A.~Ost, ``{A Multi-scale Boosted Detector for
  Efficient and Robust Gesture Recognition},'' in \emph{ECCVW}, 2014.

\bibitem{chalearn3}
J.~Y. Chang, ``{Nonparametric Gesture Labeling from Multi-modal Data},'' in
  \emph{ECCV Workshop}, 2014.

\bibitem{Nandakumar2013}
K.~Nandakumar \emph{et~al.}, ``{A Multi-modal Gesture Recognition System Using
  Audio, Video, and Skeletal Joint Data Categories and Subject Descriptors},''
  in \emph{ICMI Workshop}, 2013.

\bibitem{Le2011a}
Q.~V. Le, W.~Y. Zou, S.~Y. Yeung, and A.~Y. Ng, ``{Learning hierarchical
  invariant spatio-temporal features for action recognition with independent
  subspace analysis},'' in \emph{CVPR}, 2011.

\bibitem{Ranzato2007}
M.~Ranzato, F.~J. Huang, Y.-L. Boureau, and Y.~LeCun, ``{Unsupervised Learning
  of Invariant Feature Hierarchies with Applications to Object Recognition},''
  in \emph{CVPR}, 2007.

\bibitem{Chen2010a}
B.\hspace{2pt}Chen, J.-A.\hspace{2pt}Ting, B.\hspace{2pt}Marlin, and
  N.~de~Freitas, ``{Deep learning of invariant Spatio-Temporal Features from
  Video},'' in \emph{NIPSW}, 2010.

\bibitem{Ji2013}
S.~Ji, W.~Xu, M.~Yang, and K.~Yu, ``{3D Convolutional Neural Networks for Human
  Action Recognition},'' \emph{PAMI}, 2013.

\bibitem{chalearn5}
L.\hspace{1pt}Pigou, S.\hspace{1pt}Dieleman, and P.-J.\hspace{1pt}Kindermans,
  ``{Sign Language Recognition Using Convolutional Neural Networks},'' in
  \emph{ECCVW}, 2014.

\bibitem{chalearn6}
D.~Wu, ``{Deep Dynamic Neural Networks for Gesture Segmentation and
  Recognition},'' in \emph{ECCV Workshop}, 2014.

\bibitem{mkl2004}
F.~Bach, G.~Lanckriet, and M.~Jordan, ``{Multiple Kernel Learning, Conic
  Duality, and the SMO Algorithm},'' in \emph{ICML}, 2004.

\bibitem{gehler2009}
P.~Gehler and S.~Nowozin, ``{On Feature Combination for Multiclass Object
  Classification},'' in \emph{ICCV}, 2009.

\bibitem{ye2012}
G.~Ye, D.~Liu, I.-H. Jhuo, and S.-F. Chang, ``{Robust Late Fusion With Rank
  Minimization},'' in \emph{CVPR}, 2012.

\bibitem{natarajan2012}
P.~Natarajan, S.~Wu, S.~Vitaladevuni, X.~Zhuang, S.~Tsakalidis, U.~Park,
  R.~Prasad, and P.~Natarajan, ``{Multimodal Feature Fusion for Robust Event
  Detection in Web Videos},'' in \emph{CVPR}, 2012.

\bibitem{Ngiam2011}
J.~Ngiam, A.~Khosla, M.~Kin, J.~Nam, H.~Lee, and A.~Y. Ng, ``{Multimodal deep
  learning},'' in \emph{ICML}, 2011.

\bibitem{Srivastava2013}
N.~Srivastava and R.~Salakhutdinov, ``{Multimodal learning with Deep Boltzmann
  Machines},'' in \emph{NIPS}, 2013.

\bibitem{wu2014mm}
Z.~Wu, Y.-G. Jiang, J.~Wang, J.~Pu, and X.~Xue, ``{Exploring Inter-feature and
  Inter-class Relationships with Deep Neural Networks for Video
  Classification},'' in \emph{ACM Multimedia}, 2014.

\bibitem{Neverova2013}
N.~Neverova, C.~Wolf, G.~Paci, G.~Sommavilla, G.~W. Taylor, and F.~Nebout, ``{A
  multi-scale approach to gesture detection and recognition},'' in \emph{ICCV
  Workshop}, 2013.

\bibitem{hernandezvela2014}
A.~Hernandez-Vela \emph{et~al.}, ``{Probability-based Dynamic Time Warping and
  Bag-of-Visual-and-Depth-Words for Human Gesture Recognition in RGB-D},'' in
  \emph{Pattern Recognition Letters}, 2014.

\bibitem{neverovaECCVW2014}
N.~Neverova, C.~Wolf, G.~Taylor, and F.~Nebout, ``{Multi-scale deep learning
  for gesture detection and localization},'' in \emph{ECCVW}, 2014.

\bibitem{zanfir2013movingpose}
M.~Zanfir, M.~Leordeanu, and C.~Sminchisescu, ``{The Moving Pose: An Efficient
  3D Kinematics Descriptor for Low-Latency Action Recognition and Detection},''
  in \emph{ICCV}, 2013.

\bibitem{Deng2013}
L.\hspace{1pt}Deng, \hspace{-1pt}J.\hspace{1pt}Li,
  \hspace{-1pt}J.\hspace{1pt}Huang, \hspace{-1pt}K.\hspace{1pt}Yao,
  \hspace{-1pt}D.\hspace{1pt}Yu, \hspace{-1pt}F.\hspace{1pt}Seide,
  \hspace{-1pt}M.\hspace{1pt}Seltzer, \hspace{-1pt}G.\hspace{1pt}Zweig,
  \hspace{-1pt}X.\hspace{1pt}He, \hspace{-1pt}J.\hspace{1pt}Williams,
  \hspace{-1pt}Y.\hspace{1pt}Gong, and \hspace{-1pt}A.\hspace{1pt}Acero,
  ``{Recent advances in deep learning for speech recognition at Microsoft},''
  in \emph{\hspace{-1pt}ICASSP}, \hspace{-1pt}2013.

\bibitem{alexandre2001}
L.~A. Alexandre, A.~C. Campilho, and M.~Kamel, ``{On combining classifiers
  using sum and product rules},'' in \emph{Pattern Recognition Letters},
  no.~22, 2001, pp. 1283--1289.

\bibitem{maxout}
I.~J. Goodfellow, D.~Warde-Farley, M.~Mirza, A.~Courville, and Y.~Bengio,
  ``{Maxout Networks},'' in \emph{arXiv:1302.4389v4}, 2013.

\bibitem{BaldiSadowski20124}
P.~Baldi and P.~Sadowski, ``The dropout learning algorithm,'' \emph{Journal of
  Artificial Intelligence}, vol. 210, pp. 78--122, 2014.

\bibitem{dropout}
G.~E. Hinton, N.~Srivastava, A.~Krizhevsky, I.~Sutskever, and R.~R.
  Salakhutdinov, ``{Improving neural networks by preventing co-adaptation of
  feature detectors},'' in \emph{arXiv:1207.0580}, 2012.

\bibitem{fastdropout2013}
S.\hspace{3pt}Wang and C.\hspace{3pt}Manning, ``{Fast dropout training},'' in
  \emph{ICML}, 2013.

\bibitem{lehmann1998}
E.~L. Lehmann, ``{Elements of Large-Sample Theory},'' in \emph{ICML}, 1998.

\bibitem{fang2007}
K.~Fang, W.~Gao, and D.~Zhao, ``{Large-Vocabulary Continuous Sign Language
  Recognition Based on Transition-Movement Models},'' in \emph{IEEE
  Transactions on Systems, Man, and Cybernetics}, 2007.

\bibitem{escalera2013chalearn}
S.~Escalera \emph{et~al.}, ``{Multi-modal Gesture Recognition Challenge 2013:
  Dataset and Results},'' in \emph{ICMI workshop}, 2013.

\bibitem{Theano}
J.~Bergstra \emph{et~al.}, ``{Theano: A CPU and GPU Math Expression
  Compiler},'' in \emph{Proceedings of the Scipy Conference}, 2010.

\bibitem{geurts2006}
P.~Geurts, D.~Ernst, and L.~Wehenkel, ``{Extremely randomized trees},'' in
  \emph{Machine learning, 63(1), 3-42}, 2006.

\bibitem{bib:confinterspeechLeeKS01}
A.~Lee, T.~Kawahara, and K.~Shikano, ``Julius - an open source real-time large
  vocabulary recognition engine,'' in \emph{Interspeech}, 2001.

\bibitem{chalearn7}
N.\hspace{2pt}Camgoz, A.\hspace{2pt}Kindiroglu, and L.\hspace{2pt}Akarun,
  ``{Gesture Recognition using Template Based Random Forest Classifiers},'' in
  \emph{ECCVW}, 2014.

\bibitem{chalearn8}
G.~Evangelidis, G.~Singh, and R.~Horaud, ``{Continuous gesture recognition from
  articulated poses},'' in \emph{ECCV Workshop}, 2014.

\bibitem{chalearn4}
X.~Peng, L.~Wang, and Z.~Cai, ``{Action and Gesture Temporal Spotting with
  Super Vector Representation},'' in \emph{ECCVW}, 2014.

\bibitem{chalearn10}
G.\hspace{3pt}Chen, D.\hspace{3pt}Clarke, M.\hspace{3pt}Giuliani,
  D.\hspace{3pt}Weikersdorfer, and A.\hspace{3pt}Knoll, ``{Multi-modality
  Gesture Detection and Recognition With Un-supervision, Randomization and
  Discrimination},'' in \emph{ECCVW}, 2014.

\bibitem{mnist}
Y.~LeCun, L.~Bottou, Y.~Bengio, and P.~Haffner, ``{Gradient-based learning
  applied to document recognition},'' in \emph{Proceedings of the IEEE}, vol.
  86(11), 1998, pp. 2278--2324.

\end{thebibliography}

\ifCLASSOPTIONcaptionsoff
  \newpage
\fi

\vspace*{-17pt}
\begin{IEEEbiography}[{\includegraphics[width=1in,height=1.25in,clip,keepaspectratio]{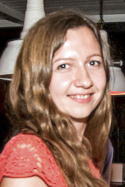}}]{Natalia Neverova}
is a PhD student at INSA de Lyon and LIRIS (CNRS, France) working in the area of gesture and action recognition with emphasis on multi-modal aspects and deep learning methods. She is advised by Christian Wolf  and Graham Taylor, and her research is a part of Interabot project in partnership with Awabot SAS. She was a visiting researcher at University of Guelph in 2014 and at Google in 2015. 
%She holds a Europeen CIMET Erasmus Mundus MSc degree with excellent distinction.\\
\vspace*{-17pt}
\end{IEEEbiography}
\begin{IEEEbiography}[{\includegraphics[width=1in,height=1.25in,clip,keepaspectratio]{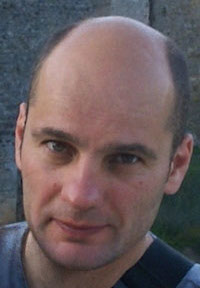}}]{Christian Wolf}
is assistant professor at INSA de Lyon and LIRIS, CNRS, since 2005. He is interested in computer vision and machine learning, especially the visual analysis of complex scenes in motion, %: gesture and activity recognition. His work puts an emphasis 
%and models of complex interactions, on
 structured models, graphical models and deep learning. 
He received his MSc in computer science from Vienna University of Technology in 2000, and a PhD from the National Institute of Applied Science (INSA de Lyon), in 2003. In 2012 he obtained the habilitation diploma.\vspace*{-17pt} 
\end{IEEEbiography}
\begin{IEEEbiography}[{\includegraphics[width=1in,height=1.25in,clip,keepaspectratio]{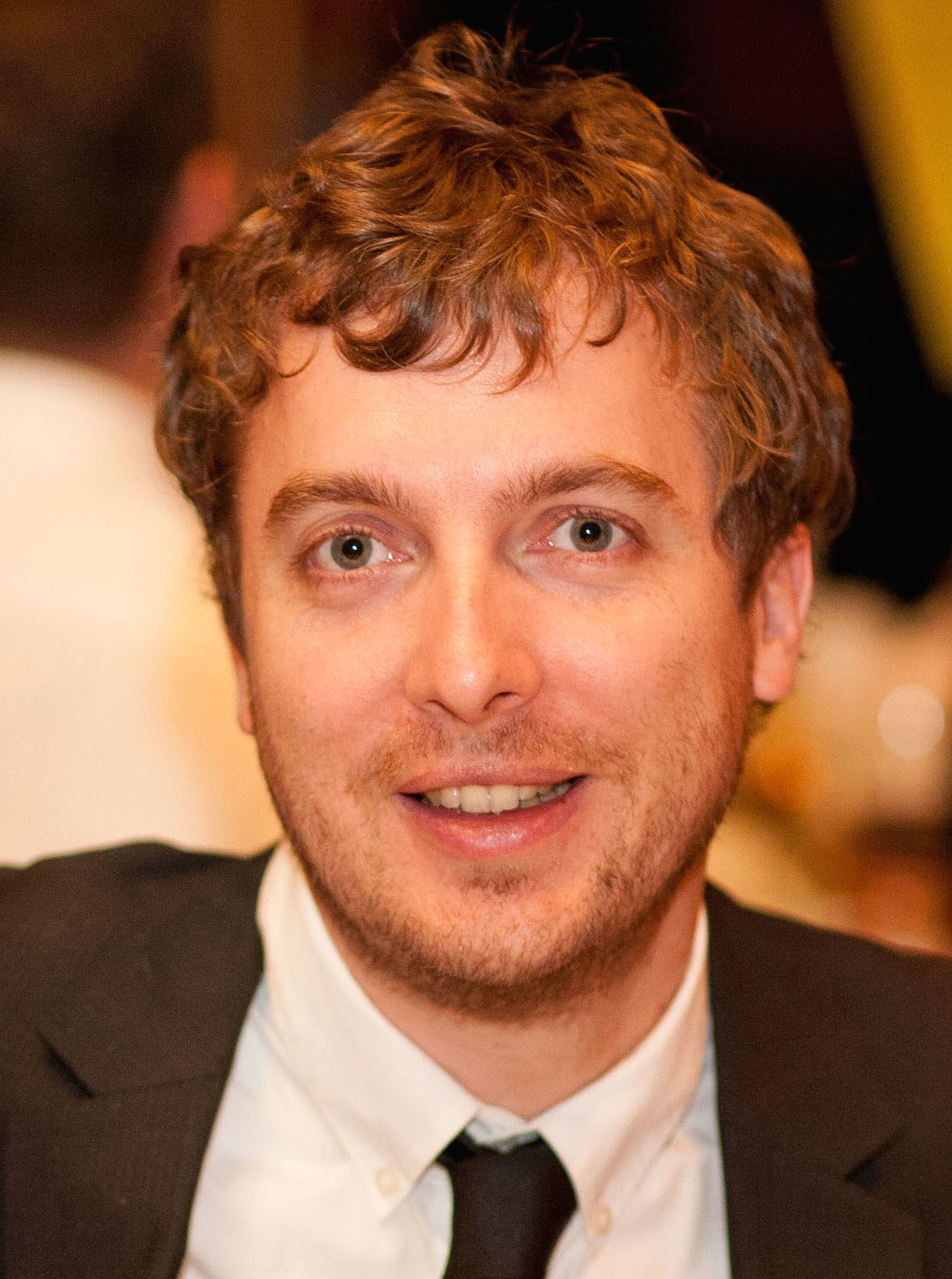}}]{Graham Taylor}
is an assistant professor at University of Guelph. He is interested in statistical machine learning and biologically-inspired computer vision, with an emphasis on unsupervised learning and time series analysis. He completed his PhD at the University of Toronto in 2009, where his thesis co-advisors were Geoffrey Hinton and Sam Roweis. He did a postdoc at NYU with Chris Bregler, Rob Fergus, and Yann LeCun.\vspace*{-17pt}
\end{IEEEbiography}
\begin{IEEEbiography}[{\includegraphics[width=1in,height=1.25in,clip,keepaspectratio]{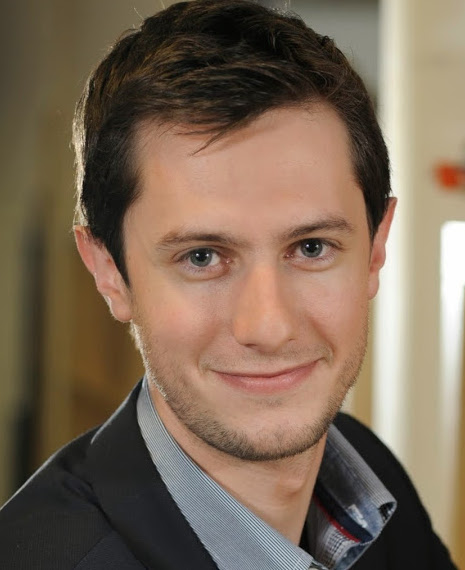}}]{Florian Nebout}
is a project manager at Awabot, leading different robotics projects, which are mostly dedicated to remote presence applications. He is especially interested in human robot interaction and how robotics can change our daily life. He graduated from INSA de Lyon, France and worked at CSIRO Human Computer Interaction team and ANU Information and Human Centred Computing Research Group in Canberra, Australia. 
\end{IEEEbiography}%\newpage

\vfill


\end{document}